%% file: main.tex
\documentclass{bmvc2k}
\usepackage{multirow}

\title{The Devil is in the Tails:\\ Fine-grained Classification in the Wild}

\addauthor{Grant Van Horn}{gvanhorn@caltech.edu}{1}
\addauthor{Pietro Perona}{perona@caltech.edu}{1}

\addinstitution{
 Caltech
}

\runninghead{Van Horn, Perona}{The Devil is in the Tails}

\def\eg{\emph{e.g}\bmvaOneDot}

\def\etal{\emph{et al}\bmvaOneDot}
\def\ie{i.e\bmvaOneDot}

\newif\ifsqueeze
\squeezetrue  
\ifsqueeze
  \newcommand{\Caption}[1]{\caption{{\footnotesize #1}}}

\else
  \newcommand{\Caption}[1]{\caption{#1}}

\fi

\begin{document}

\maketitle

\begin{abstract}
The world is long-tailed. What does this mean for computer vision and visual recognition? The main two implications are (1)  the number of categories we need to consider in applications can be  very large, and (2) the number of training examples for most categories can be very small. Current visual recognition algorithms have achieved excellent classification  accuracy. However, they require many training examples to reach peak performance, which suggests that long-tailed distributions will not be dealt with well. We analyze this question in the context of eBird, a large fine-grained classification dataset, and a state-of-the-art deep network classification algorithm.  We find that (a) peak classification performance on well-represented categories is excellent, (b) given enough data, classification performance suffers only minimally from an increase in the number of classes, (c) classification performance decays precipitously as the number of training examples decreases, (d) surprisingly, transfer learning is virtually absent in current methods. Our findings suggest that our community should come to grips with  the question of long tails. 
\end{abstract}

\input{introduction}

\input{related_work}

\input{setup}

\input{experiments}

\input{conclusion}
\clearpage

\input{supplementary}
\clearpage

\bibliography{main}
\end{document}

%% file: introduction.tex
\section{Introduction}
\label{sec:intro}


\begin{figure}
\centering
\subfigure[]{\label{fig:real_world_long_tail}\includegraphics[width=0.32\textwidth]{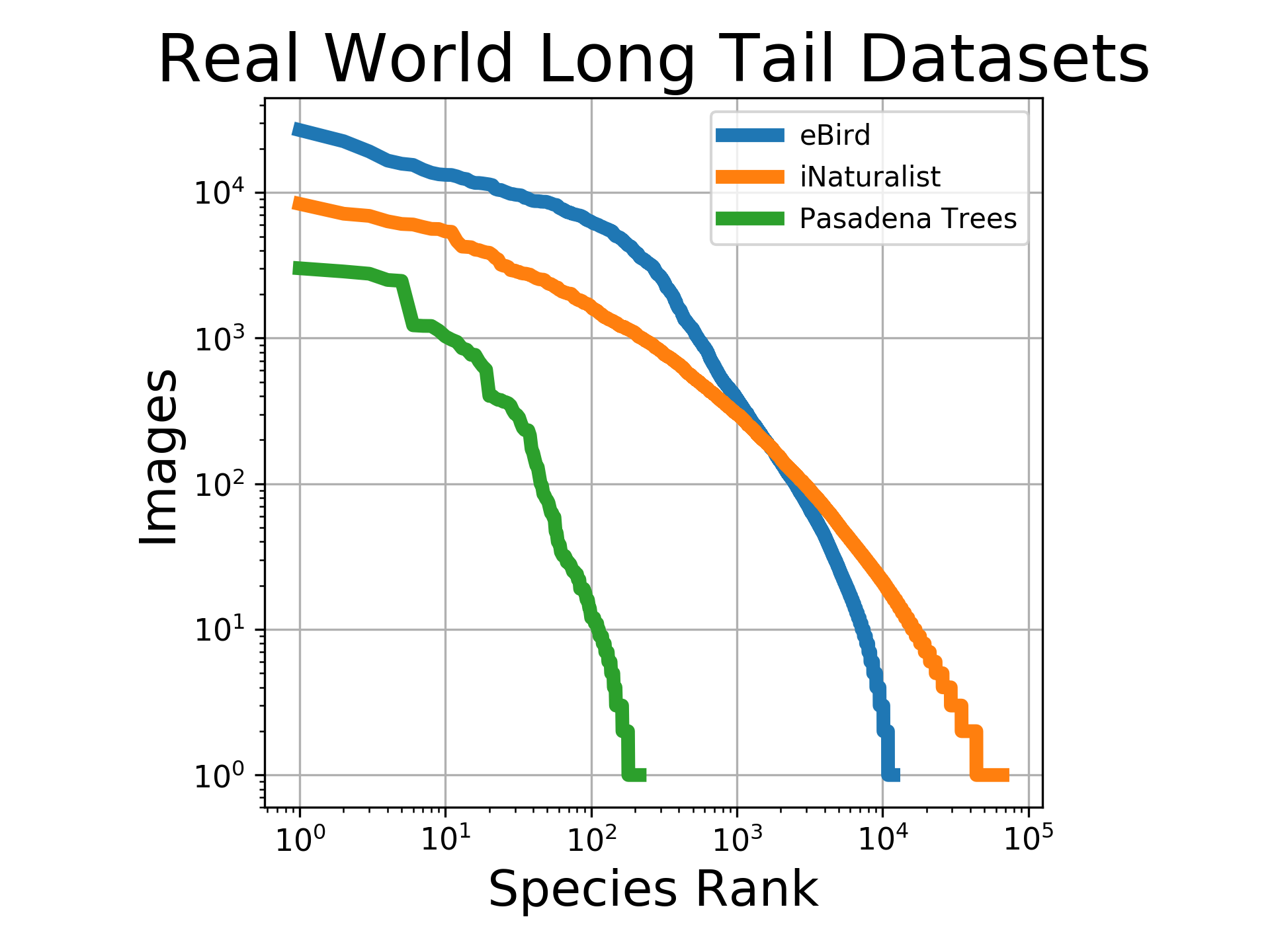}}
\subfigure[]{\label{fig:experimental_long_tail}\includegraphics[width=0.32\textwidth]{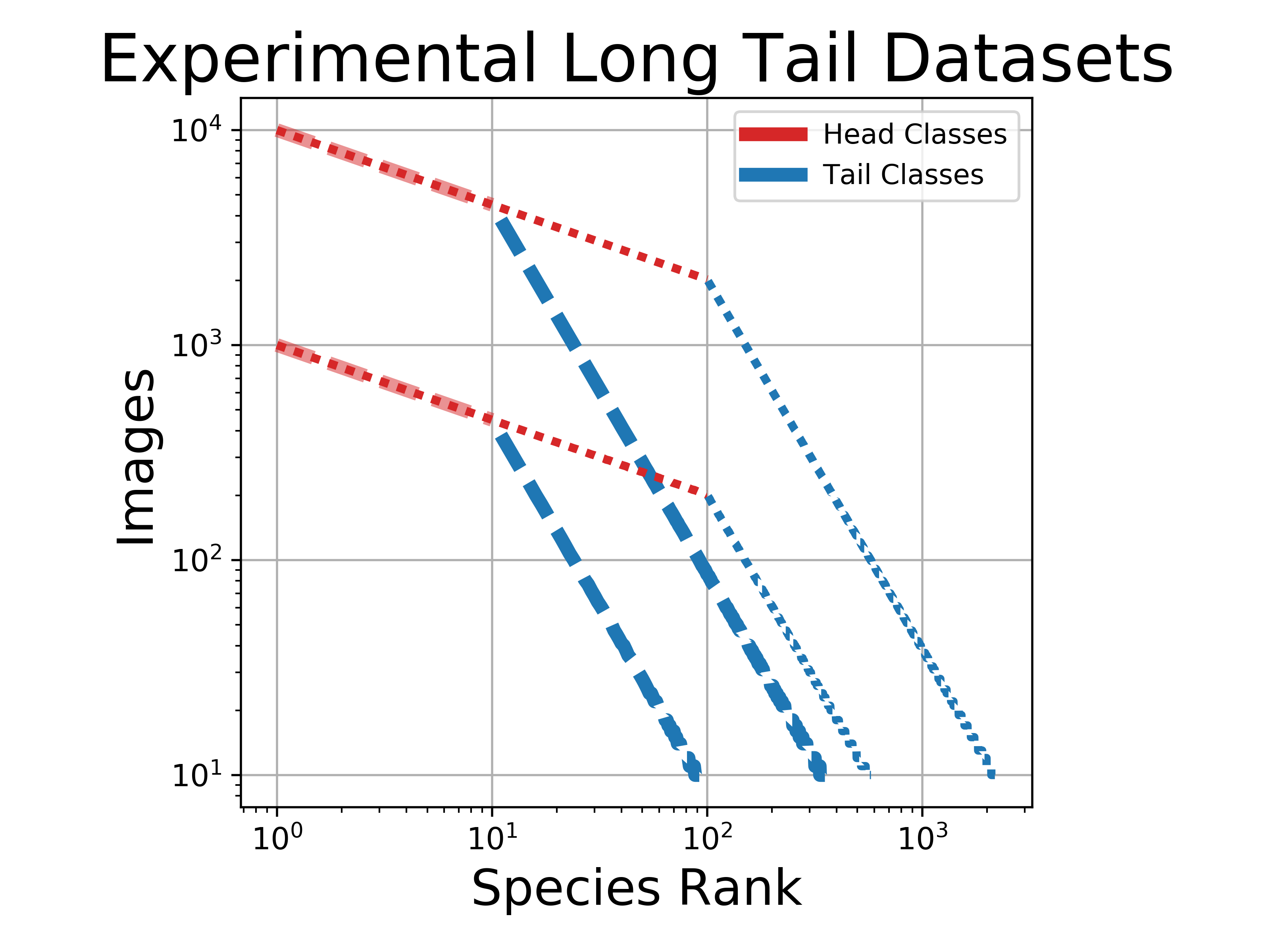}}
\subfigure[]{\label{fig:long_tail_vs_approx}\includegraphics[width=0.32\textwidth]{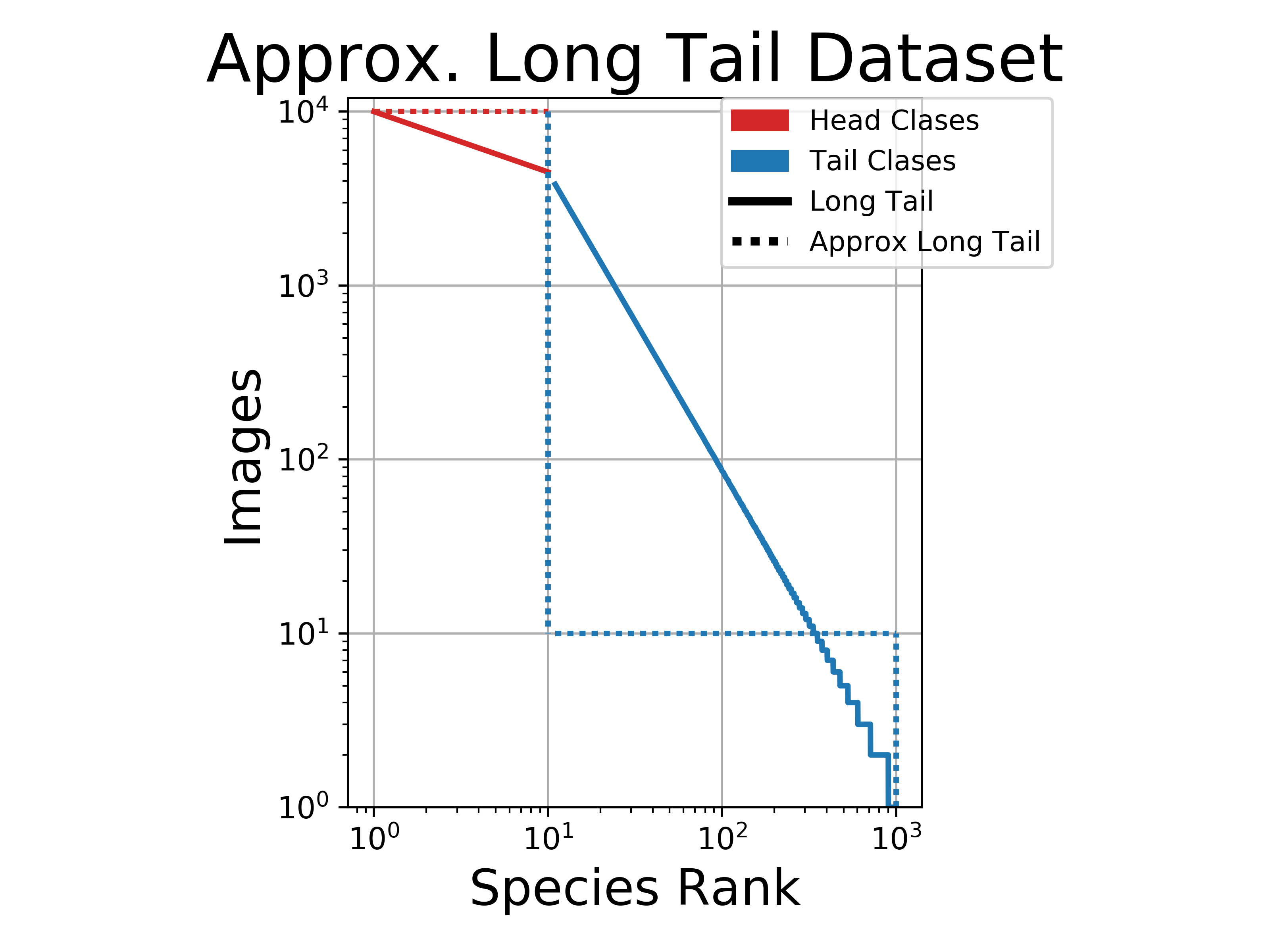}}
\Caption{{\bf (a) The world is long-tailed.} Class frequency statistics in real world datasets (birds, a wide array of natural species, and trees). These are long-tailed distributions where a few classes have many examples and most classes have few. {\bf (b) The 4 experimental long tail datasets used in this work.} We modeled the eBird dataset (blue curve in \textbf{(a)}) and created four long tail datasets by shifting the modeled eBird dataset down (fewer images) and to the left (fewer species) by different amounts. Classes are split into  head and tail groups; images per class in the respective groups decay exponentially. {\bf (c) Approximation of a long tail dataset.} This approximation allows us to more easily study the effects of head classes on tail class performance.}
\label{fig:real_experimental_datasets}
\end{figure}

During the past five years we have witnessed dramatic improvement in the performance of visual recognition algorithms~\cite{russakovsky2015imagenet}. Human performance has been approached or achieved in many instances. Three concurrent developments have enabled such progress: (a) the invention of `deep network' algorithms where visual computation is learned from the data rather than hand-crafted by experts~\cite{fukushima1982neocognitron,lecun1989backpropagation,krizhevsky2012imagenet}, (b) the design and construction of large and well annotated datasets~\cite{feifeiFP04,everinghamEtal05,deng2009imagenet,lin2014microsoft} supplying researchers with a sufficient amount of data to train learning-based algorithms, and (c) the availability of inexpensive and ever more powerful computers, such as GPUs~\cite{lindholm2008nvidia}, for algorithm training. 

Large annotated datasets yield two additional benefits, besides supplying  deep nets with sufficient training fodder. The first is offering common performance benchmarks that allow researchers to compare results and quickly evolve the best algorithmic architectures. The second, more subtle but no less important, is providing researchers with a compass  -- a definition of the visual tasks that one ought to try and solve. Each new dataset pushes us a bit further towards solving real world challenges. We wish to focus here on the latter aspect.

One goal of visual recognition is to enable machines to recognize objects in the world. What does the world look like? In order to better understand the nature of visual categorization in the wild we examined three real-world datasets: bird species, as photographed worldwide by birders who are members of eBird~\cite{sullivan2009ebird}, tree species, as observed along the streets of Pasadena~\cite{wegner2016cataloging}, and plants and animal species, as photographed by the iNaturalist (www.inaturalist.org) community. One salient aspect of these datasets is that some species are very frequent, while most species are represented by only few specimens (Fig~\ref{fig:real_world_long_tail}). In a nutshell: the world is long-tailed, as previously noted in the context of subcategories and object views~\cite{salakhutdinov2011learning,zhu2014capturing}. This is in stark contrast with current datasets for visual classification, where specimen distribution per category is almost uniformly distributed (see~\cite{lin2014microsoft} Fig. 5(a)). 

With this observation in mind, we ask whether current state-of-the-art classification algorithms, whose development is motivated and benchmarked by uniformly distributed datasets,  deal well with the world's long tails. Humans appear to be able to generalize from few examples, can our algorithms do the same? Our experiments show that the answer is  {\em no}. While, when data is abundant, machine vision classification performance can currently rival humans, we find that this is emphatically not the case when data is scarce for most classes, even if a few are abundant.

This work is organized in the following: In Section~\ref{sec:related} we review the related work. We then describe the datasets and training process in Section~\ref{sec:setup} followed by an analysis of the experiments in Section~\ref{sec:experiments}. We summarize and conclude in Section~\ref{sec:conclusion}.

%% file: related_work.tex
\section{Related Work}
\label{sec:related}







\textbf{Fine-Grained Visual Classification} -- The vision community has released many fine-grained datasets covering several domains such as birds \cite{welinder2010caltech,wah2011caltech,berg2014birdsnap,van2015building}, dogs \cite{KhoslaYaoJayadevaprakashFeiFei_FGVC2011,liu2012dog}, airplanes \cite{maji2013fine,vedaldi2014understanding}, flowers \cite{nilsback2006visual}, leaves \cite{kumar2012leafsnap}, trees \cite{wegner2016cataloging} and cars\cite{krause20133d,lin2014jointly}. These datasets were constructed to be uniform, or to contain "enough" data for the task. The recent Pasadena Trees dataset \cite{wegner2016cataloging} is the exception. Most fine-grained research papers present a novel model for classification \cite{xu2015augmenting,lin2015bilinear,farrell2011birdlets,krause2015fine,xie2015hyper,branson2014improved,gavves2015local,simon2015neural,goring2014nonparametric,shih2015part,zhang2014part,berg2013poof,chai2013symbiotic,xiao2015application,zhang2016weakly,pu2014looks}. While these methods often achieve state-of-the-art performance at the time of being published, it is often the case that the next generation of convolutional networks can attain the same level of performance without any custom modifications. In this work we use the Inception-v3 model~\cite{szegedy2016rethinking}, pretrained on ImageNet for our experiments. Some of the recent fine-grained papers have investigated augmenting the original datasets with additional data from the web \cite{krause2016unreasonable,xu2015augmenting,xie2015hyper,van2015building}. Krause \etal \cite{krause2016unreasonable} investigated the collection and use of a large, noisy dataset for the task of fine-grained classification and found that off the shelf CNN models can readily take advantage of these datasets to increase accuracy and reach state-of-the-art performance. Krause \etal mention, but do not investigate, the role of the long tail distribution of training images. In this work we specifically investigate the effect of this long tail on the model performance.


\textbf{Imbalanced Datasets } -- Techniques to handle imbalanced datasets are typically split into two regimes: algorithmic solutions and data solutions. In the first regime, cost-sensitive learning~\cite{elkan2001foundations} is employed to force the model to adjust its decision boundaries by incurring a non-uniform cost per misclassification; see~\cite{he2009learning} for a review of the techniques. The second regime concerns data augmentation, achieved either through over-sampling the minority classes, under sampling the majority classes or synthetically generating new examples for the minority classes. When using mini batch gradient descent (as we do in the experiments), oversampling the minority classes is similar to weighting these classes more than the majority classes, as in cost-sensitive learning. We conduct experiments on over-sampling the minority classes. We also employ affine~\cite{krizhevsky2012imagenet} and photometric~\cite{howard2013some} transformations to synthetically boost the number of training examples.  




 
\textbf{Transfer Learning} -- Transfer learning~\cite{pan2010survey} attempts to adapt the representations learned in one domain to another. In the era of deep networks, the simplest form of transfer learning is using features extracted from pretrained ImageNet~\cite{russakovsky2015imagenet} or Places~\cite{zhou2014learning} networks, see \cite{sharif2014cnn, donahue2014decaf}. The next step is actually fine-tuning~\cite{girshick2014rich} these pretrained networks for the target task \cite{yosinski2014transferable,agrawal2014analyzing,oquab2014learning,huh2016makes}. This has become the standard method for obtaining baseline numbers on a new target dataset and often leads to impressive results~\cite{azizpour2015generic}, especially when the target dataset has sufficient training examples. More sophisticated transfer learning methods~\cite{long2015learning, tzeng2015simultaneous} are aimed at solving the domain adaptation problem. In this work, we are specifically interested in a single domain, which happens to contain a long tail distribution of training data for each class. We investigate whether there is a transfer of knowledge from the well represented classes to the sparsely represented classes. 

\textbf{Low Shot Learning} -- We experiment with a minimum of 10 training images per class, which falls into the realm of low shot learning, a field concerned with learning novel concepts from few examples. In \cite{wang2016learning} Wang and Herbet learn a regression function from classifiers trained on small datasets to classifiers trained on large datasets, using a fixed feature representation. Our setup is different in that we want to allow our feature representation to adapt to the target dataset and we want a model that can classify both the well represented classes and the sparsely represented classes. The recent work of Hariharan and Girshick in \cite{hariharanlow} explored this setup specifically, albeit in the broad domain of ImageNet. The authors propose a low shot learning benchmark and implement a loss function and feature synthesis scheme to boost performance on under represented classes. However, their results showed marginal improvement when using a high capacity model (at 10 images per class the ResNet-50 \cite{he2016deep} model performed nearly as well as their proposed method). Our work aims to study the relationship between the well represented classes and the sparse classes, within a single domain. Metric learning tackles the low shot learning problem by learning a representation space where distance corresponds to similarity. While these techniques appear promising and provide benefits beyond classification, they do not hold up well against simple baseline networks for the specific task of classification \cite{rippel2015metric}.

%% file: setup.tex
\section{Experiment Setup}
\label{sec:setup}


\subsection{Datasets}
\label{sec:dataset_desc}

We consider three different types of datasets: uniform, long tail and approximate long tail. We used images from eBird (ebird.org) to construct each of these datasets. These images are real world observations of birds captured by citizen scientists and curated by regional experts. Each dataset consists of a training, validation, and test split. When placing images into each split we ensure that a photographer's images do not end up in multiple splits for a single species. The test set is constructed to contain as many different photographers as possible (\eg 30 images from 30 different photographers). The validation set is similarly constructed and the train set is constructed from the remaining photographers. 

\textbf{Uniform Datasets} -- The uniform datasets allow us to study the performance of the classification model under optimal image distribution conditions. These datasets have the same number of images per class: either 10, 100, 1K, or 10K. The total number of classes can be either 10, 100, or 1K. We we did not analyze a uniform dataset with 1K classes containing 1K or 10K images each due to a lack of data from eBird. Each smaller dataset is completely contained within the larger dataset (\eg the 10 class datasets are contained within the 100 class datasets, etc.). The test and validation sets are uniform, with 30 and 10 images for each class respectively, and remain fixed for a class across all uniform datasets.  

\textbf{Approx. Long Tail Datasets} -- To conveniently explore the effect of moving from a uniform dataset to a long tail dataset we constructed approximate long tail datasets, see Figure \ref{fig:long_tail_vs_approx}. These datasets consist of 1K classes split into two groups: the head classes and the tail classes. All classes within a group have the same number of images. We study two different sized splits: a 10 head, 990 tail split and a 100 head, 900 tail split. The 10 head split can have 10, 100, 1K, or 10K images in each head class. The 100 head split can have 10, 100, or 1K images in each head class. The tail classes from both splits can have 10 or 100 images. We use the validation and test sets from the 1K class uniform dataset for all of the approximate long tail datasets. This allows us to compare the performance characteristics of the different datasets in a reliable way, and we can use the 1K class uniform datasets as reference points. 

\textbf{Long Tail Datasets} -- The full eBird dataset, with nearly 3 million images, is not amenable to easy experimentation. Rather than training on the full dataset we would prefer to model the image distribution and use it to construct smaller, tunable datasets, see Figure \ref{fig:experimental_long_tail}. We did this by fitting a two piece broken power law to the eBird image distribution. Each class $i \in [1, N]$, is put into the head group if $i <= h$ otherwise it is put into the tail group, where $h$ is the number of head classes. Each head class $i$ contains $y\cdot{}i^{a_1}$ images, where $y$ is the number of images in the most populous class and $a_1$ is the power law exponent for the head classes. Each tail class $i$ has $y\cdot{}h^{(a_1 - a_2)}\cdot{}i^{a_2}$ where $a_2$ is the power law exponent for the tail classes. We used linear regression to determine that $a_1 = -0.3472$ and $a_2=-1.7135$. We fixed the minimum number of images for a class to be 10. This leaves us with 2 parameters that we can vary: $y$, which shifts the distribution up and down, and $h$ which shifts the distribution left and right. We analyze four long tail datasets by selecting $y$ from \{1K, 10K\} and $h$ from \{10, 100\}. Each resulting dataset consists of a different number of classes and therefore has a different test and validation split. We keep to the pattern of reserving 30 test images and 10 validation images for each class.   


\subsection{ Model Training \& Testing Details}
\label{sec:train_test_details}

\textbf{Model} -- We use the Inception-v3 network\cite{szegedy2016rethinking}, pretrained from ILSVC 2012 as the starting point for all experiments. The Inception-v3 model exhibits good trade-off between size of the model (27.1M parameters) and classification accuracy on the ILSVC (78\% top 1 accuracy) as compared to architectures like AlexNet and VGG. We could have used the ResNet-50 model but opted for Inception-v3 as it is currently being used by the eBird development team. 

\textbf{Training} -- We have a fixed training regime for each dataset. We fine-tune the pretrained Inception-v3 network (using TensorFlow\cite{tensorflow2015-whitepaper}) by training all layers using a batch size of 32. Unless noted otherwise, batches are constructed by randomly sampling from the pool of all training images. The initial learning rate is 0.0045 and is decayed exponentially by a factor or 0.94 every 4 epochs. Training augmentation consists of taking random crops from the image whose area can range from 10\% to 100\% of the image, and whose aspect ratio can range from 0.7 to 1.33. The crop is randomly flipped and has random random brightness and saturation distortions applied. 

\textbf{Testing} -- We use the validation loss to stop the training by waiting for it to steadily increase, signaling that the model is overfitting.  We then consider all models up to this stopping point and use the model with the highest validation accuracy for testing. At test time, we take a center crop of the image, covering 87.5\% of the image area. We track top 1 image accuracy as the metric, as is typically used in fine-grained classification benchmarks. Note that image accuracy is the same as class average accuracy for datasets with uniform validation and test sets, as is the case for all of our experiments.

%% file: experiments.tex
\section{Experiments}
\label{sec:experiments}


\begin{figure}[h]
\centering
\subfigure[]{\label{fig:nn_svm_error}\includegraphics[width=0.45\textwidth]{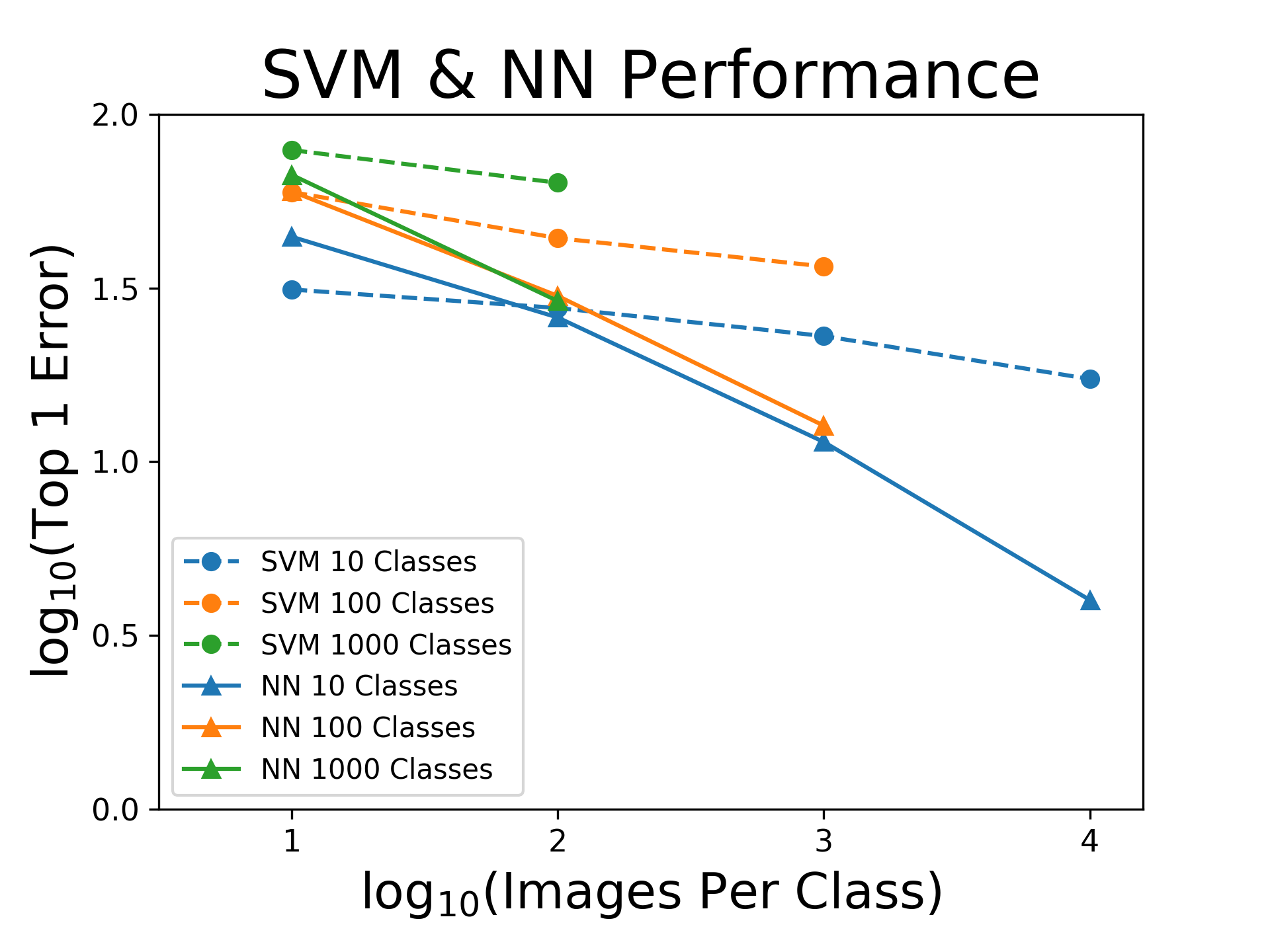}}
\subfigure[]{\label{fig:example_misclassifications}\includegraphics[width=0.45\textwidth]{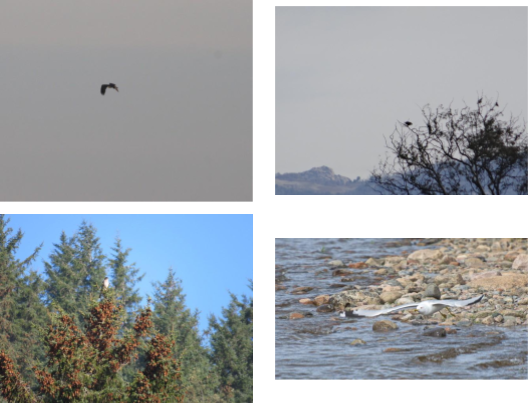}}
\Caption{{\bf (a) Classification performance as a function of training set size on uniform datasets.} A neural network (solid lines) achieves excellent accuracy on these uniform datasets. Performance keeps improving as the number of training examples increases to 10K per class --  each 10x increase in dataset size is rewarded with a 2x cut in the error rate. We also see that the neural net scales extremely well with increased number of classes, increasing error only marginally when 10x more classes are used. Neural net performance is also compared with SVM (dashed lines) trained on extracted ImageNet features.  We see that fine-tuning the neural network is beneficial in all cases except in the extreme case of 10 classes with 10 images each. {\bf (b) Example misclassifications.} Four of the twelve images misclassified by the 10 class, 10K images per class model. Clockwise from top left: Osprey misclassified as Great Blue Heron, Bald Eagle (center of image) misclassified as Great Blue Heron, Cooper's Hawk misclassified as Great Egret, and Ring-billed Gull misclassified as Great Egret.}
\label{fig:nn_svm_error_and_long_tail}
\end{figure}

\subsection{Uniform Datasets}
\label{sec:uniform_dataset_exp}

We first study the performance characteristics of the uniform datasets. We consider two regimes: 1) we extract feature vectors from the pretrained network and train a linear SVM. 2) We fine-tune the pretrained network, see section \ref{sec:train_test_details} for the training protocol. We use the activations of the layer before the final fully connected layer as our features for the SVM, and used the validation set to tune the penalty parameter. Figure \ref{fig:nn_svm_error} plots the error achieved under these two different regimes. We can see that fine-tuning the neural network is beneficial in all cases except the extreme case of 10 classes with 10 images each (in which case the model overfit quickly, even with extensive hyperparameter sweeps). The neural network scales incredibly well with increasing number of classes, incurring a small increase in error for 10x increase in the number of classes. This should be expected given that the network was designed for 1000-way ImageNet classification. At 10k images per class the network is achieving 96\% accuracy on 10 bird species, showing that the network can achieve high performance given enough data. For the network, a 10x increase in data corresponds to at least a 2x error reduction. Keep in mind that the opposite is true as well: as we remove 10x images the error rate increases by at least 2x. These uniform dataset results will be used as reference points for the long tail experiments. 


\begin{figure}[h]
\centering
\subfigure[]{\label{fig:a}\includegraphics[width=0.3\textwidth]{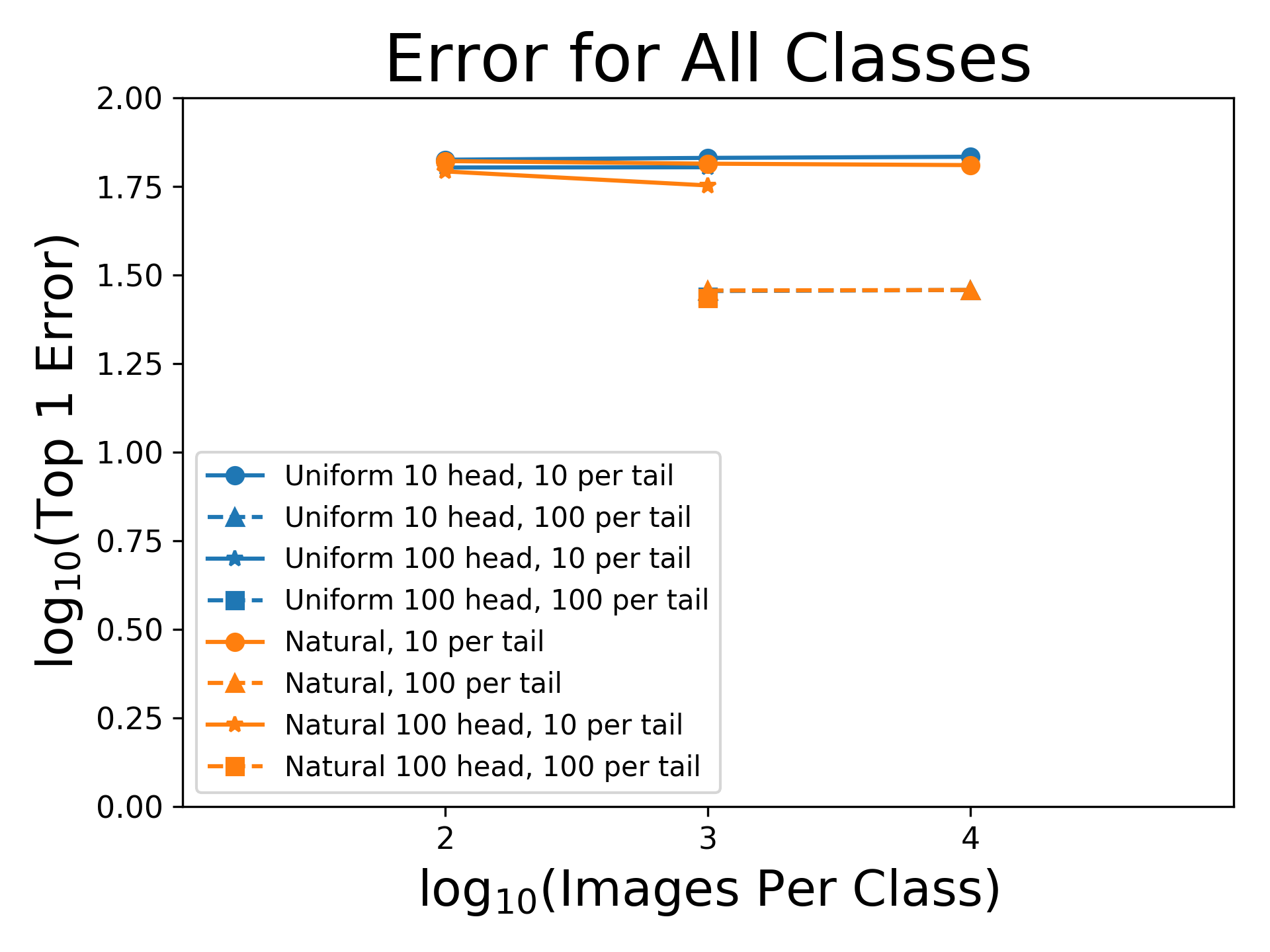}}
\subfigure[]{\label{fig:b}\includegraphics[width=0.3\textwidth]{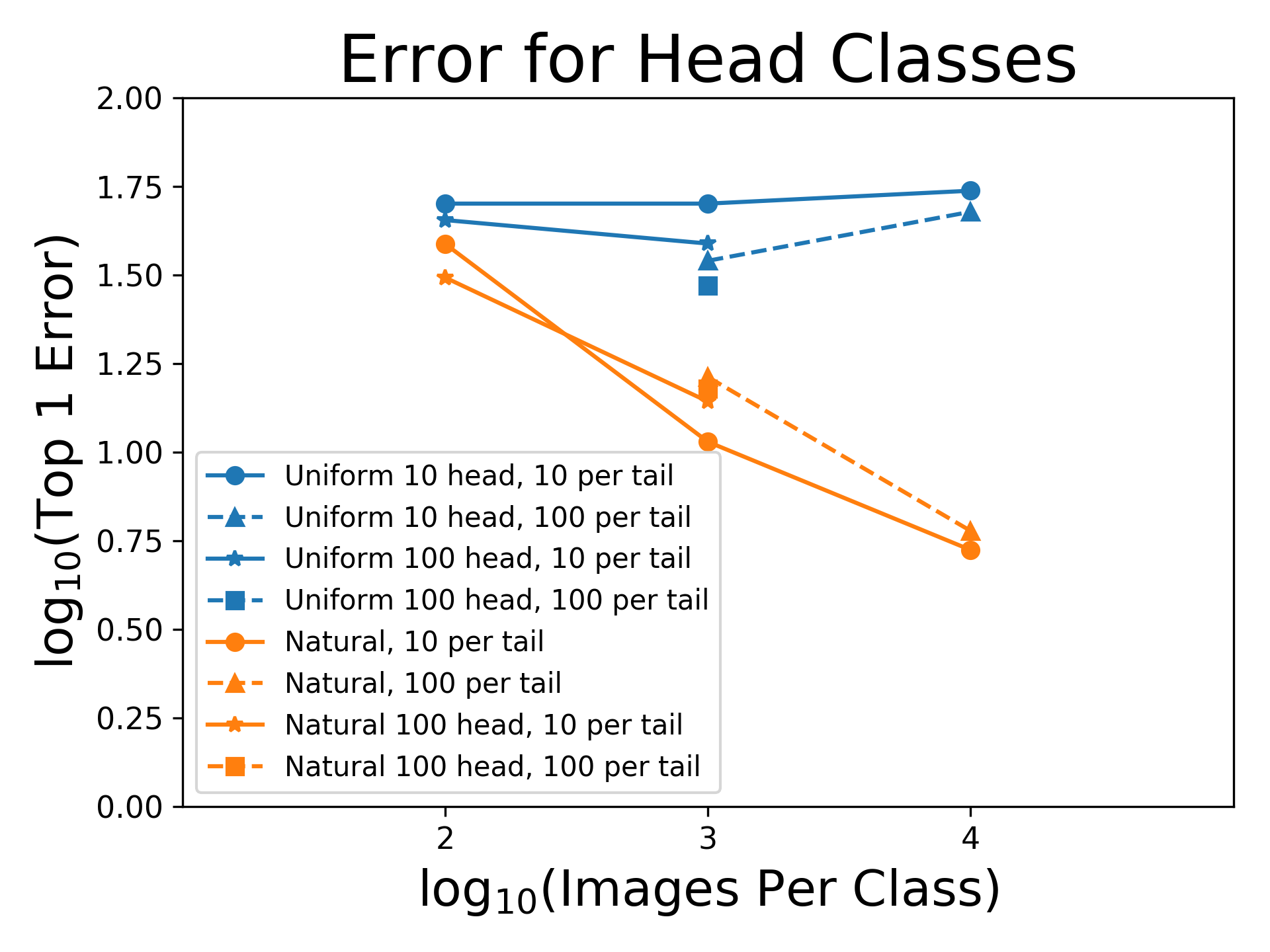}}
\subfigure[]{\label{fig:c}\includegraphics[width=0.3\textwidth]{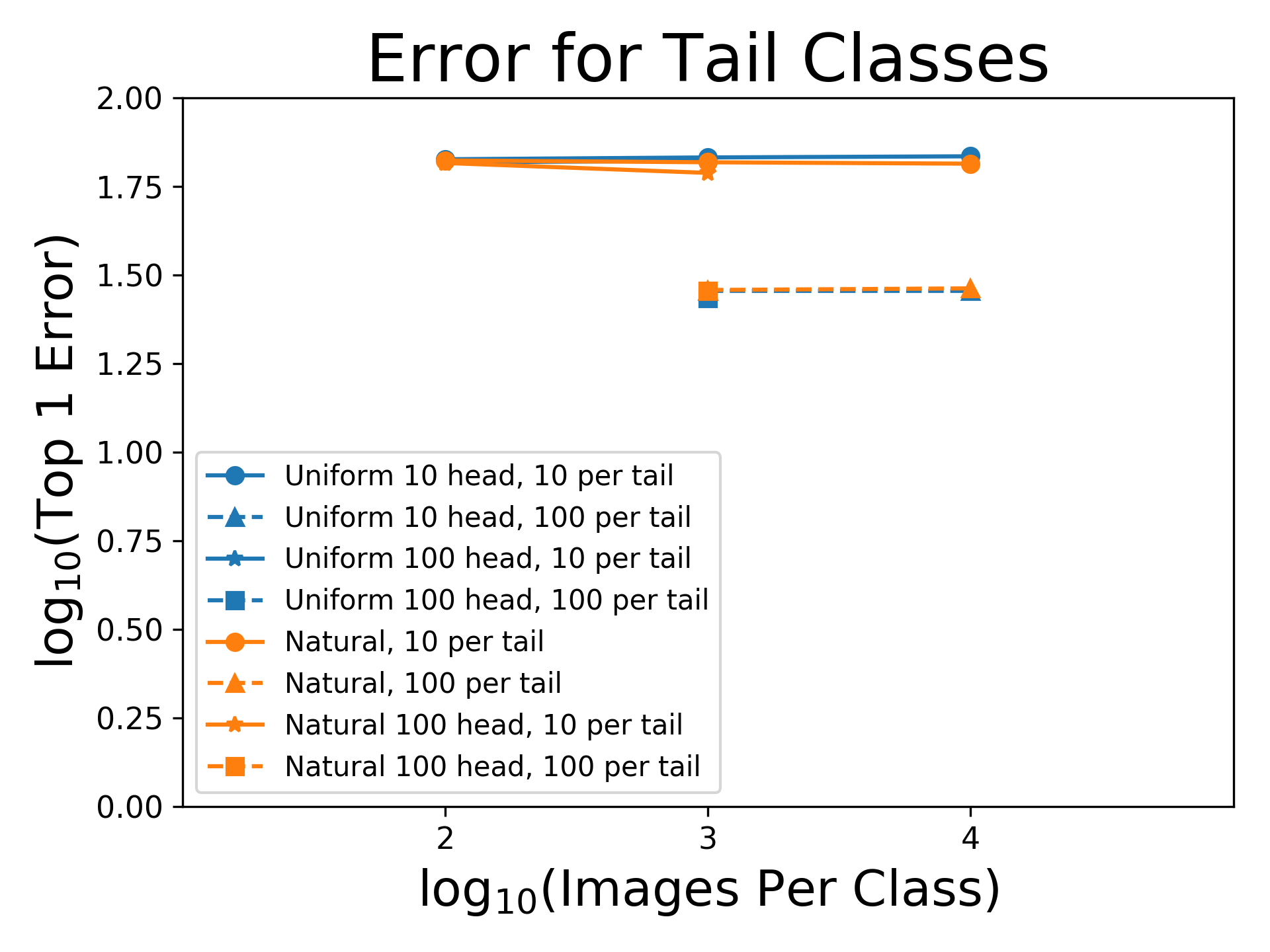}}
\Caption{{\bf Uniform vs. Natural Sampling -- effect on error.} Error plots for models trained with uniform sampling and natural sampling. \textbf{(a)} The overall error of both methods is roughly equivalent, with natural sampling tending to be as good or better than uniform sampling. \textbf{(b)} Head classes clearly benefit from natural sampling. \textbf{(c)} Tail classes tend to have the same error under both sampling regimes.}
\label{fig:uniform_vs_natural_error}
\end{figure}

\subsection{Uniform vs Natural Sampling}
\label{sec:uniform_vs_natural_sampling}

\begin{figure}[h]
\centering
\subfigure[10 Head, 990 Tail Classes]{\label{fig:a}\includegraphics[width=0.45\textwidth]{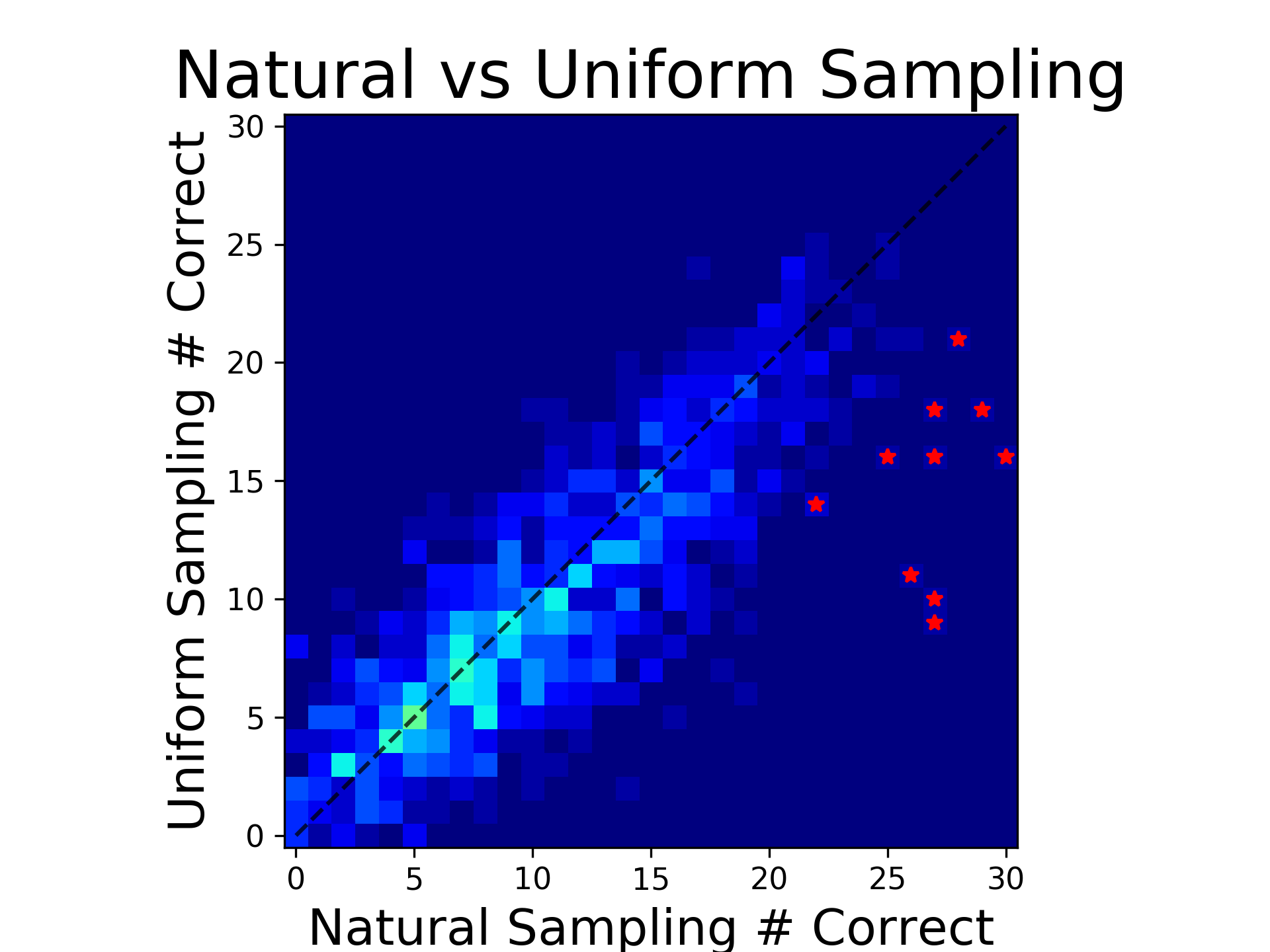}}
\subfigure[100 Head, 900 Tail Classes]{\label{fig:b}\includegraphics[width=0.45\textwidth]{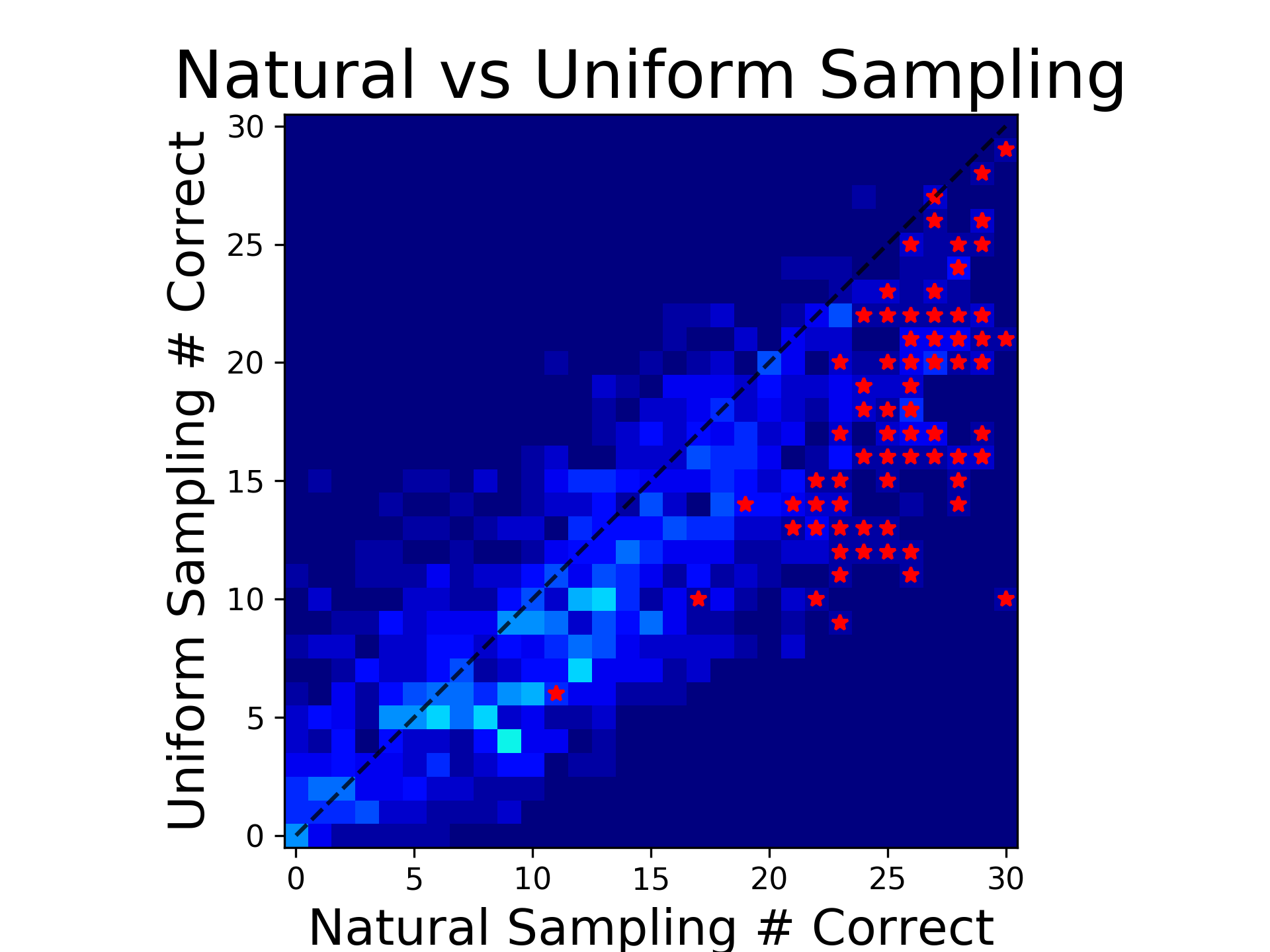}}
\Caption{{\bf Uniform vs. Natural Sampling -- effect on accuracy.} We compare the effect of uniformly sampling from classes vs sampling from their natural image distribution when creating training batches for long tailed datasets, Section \ref{sec:uniform_vs_natural_sampling}. We use 30 test images per class, so correct classification rate is binned into 31 bins. It is clear that the head classes (marked as stars) benefit from the natural sampling in both datasets. The tail classes in \textbf{(a)} have an average accuracy of 32.1\% and 34.2\% for uniform and natural sampling respectively. The tail classes in \textbf{(b)} have an average accuracy of 33.5\% and 38.6\% for uniform and natural sampling respectively. For both plots, head classes have 1000 images and tail classes have 10 images.}
\label{fig:uniform_vs_natural}
\end{figure}

The long tail datasets present an interesting question when it comes to creating the training batches: Should we construct batches of images such that they are sampled uniformly from all classes, or such that they are sampled from the natural distribution of images? Uniformly sampling from the classes will result in a given tail image appearing more frequently in the training batches than a given head image, \ie we are oversampling the tail classes. To answer this question, we trained a model for each of our approximate long tail datasets using both sampling methods and compared the results. Figure \ref{fig:uniform_vs_natural_error} plots the error achieved with the different sampling techniques on three different splits of the classes (all classes, the head classes, and the tail classes). We see that both sampling methods often converge to the same error, but the model trained with natural sampling is typically as good or better than the model trained with uniform sampling. Figure \ref{fig:uniform_vs_natural} visualizes the performance of the classes under the two different sampling techniques for two different long tail datasets. These figures highlight that the head classes clearly benefit from natural sampling and the center of mass of the tail classes is skewed slightly towards the natural sampling. The results for the long tail dataset experiments in the following sections use natural sampling.

\begin{figure}[h]
\centering
\subfigure[]{\label{fig:tail_vs_head_acc_approx}\includegraphics[width=0.45\textwidth]{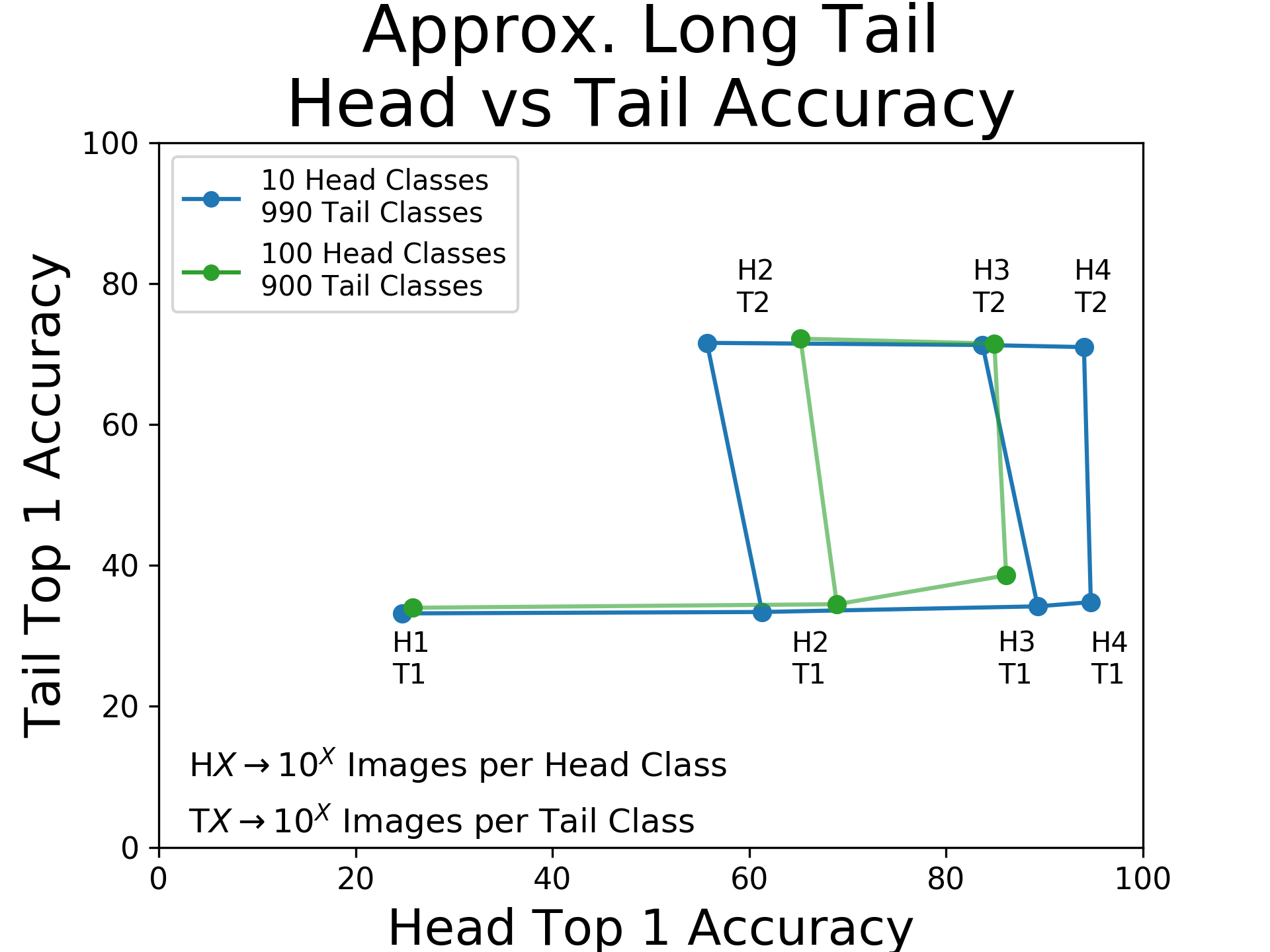}}
\subfigure[]{\label{fig:tail_vs_head_acc_long_tail}\includegraphics[width=0.45\textwidth]{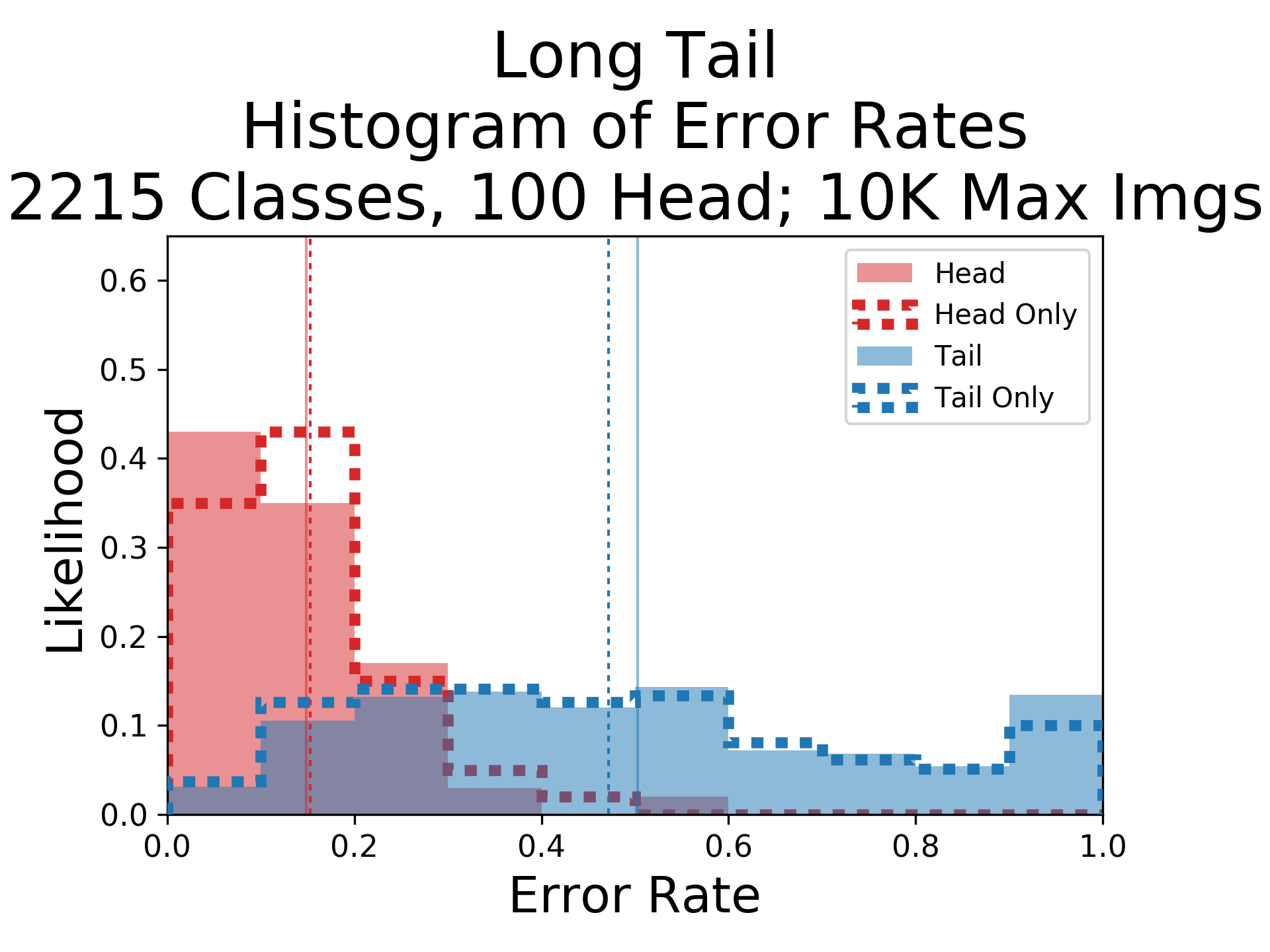}}
\Caption{{\bf Transfer between head and tail in approximate long tail datasets.} (a) Head class accuracy is plotted against tail class accuracy as we vary the number of training examples in the head and in the tail for the approximate long tail datasets. Each point is associated with its nearest label. The labels indicate (in base 10) how much training data was in each head class (H) and each tail class (T). Lines between points indicate an increase in either images per head class, or images per tail class. As we increase images in the head class by factors of 10, the performance on the tail classes remains approximately constant. This means that there is a very poor transfer of knowledge from the head classes to the tail classes. As we increase the images per tail class, we see a slight loss in performance in the head classes. The overall accuracy of the model is vastly improved though. \textbf{(b) Histogram of error rates for a long tail dataset}. The same story applies here: the tail classes do not benefit from the head classes. The overall error of the joint head and tail model is 48.6\%. See Figure \ref{fig:histogram_of_error_rates} for additional details.}
\label{fig:tail_vs_head_acc}
\end{figure}

\subsection{Transferring Knowledge from the Head to the Tail}
\label{sec:tail_performance}

Section \ref{sec:uniform_dataset_exp} showed that the Inception-v3 architecture does extremely well on uniform datasets: achieving 96\% accuracy on the 10 class, 10K images per class dataset, 87.3\% accuracy on the 100 class, 1K images per class dataset, and 71.5\% accuracy on the 1K class, 100 images per class dataset. The question we seek to answer is: How is performance affected when we move to a long tail dataset? Figure \ref{fig:tail_vs_head_acc_approx} summarizes our findings for the approximate long tail datasets. Starting with a dataset of 1000 classes and 10 images in each class, the top 1 accuracy across all classes is 33.2\% (this is the bottom, left most blue point in the figure). If we designate 10 of the classes as head classes, and 990 classes as tail classes, what happens when we increase the number of available training data in the head classes (traversing the bottom blue line in Figure \ref{fig:tail_vs_head_acc_approx})? We see that the head class accuracy approaches the peak 10 class performance of 96\% accuracy (reaching 94.7\%), while the tail classes have remained near their initial performance of 33.2\%.

We see a similar phenomena even if we are more optimistic regarding the number of available training images in the tail classes, using 100 rather than 10 (the top blue line in Figure \ref{fig:tail_vs_head_acc_approx}). The starting accuracy across all 1000 classes, each with 100 training images, is 71.5\%. As additional images are added to the head classes, the accuracy on the head classes again approaches the peak 10 class performance (reaching 94\%) while the tail classes are stuck at 71\%. 

We can be optimistic in another way by moving more classes into the head, therefore making the tail smaller. We now specify 100 classes to be in the head, leaving 900 classes for the tail (the green points in \ref{fig:tail_vs_head_acc_approx}). We see a similar phenomena even in this case, although we do see a slight improvement for the tail classes when the 100 head classes have 1k images each. These experiments reveal that there is very little to no transfer learning occurring within the network. Only the classes that receive additional training data actually see an improvement in performance, even though all classes come from the same domain. To put it plainly, an additional 10K bird images covering 10 bird species does nothing to help the network learn a better representation for the remaining bird species.


\begin{figure}
\centering
\subfigure[]{\label{fig:a}\includegraphics[width=0.3\textwidth]{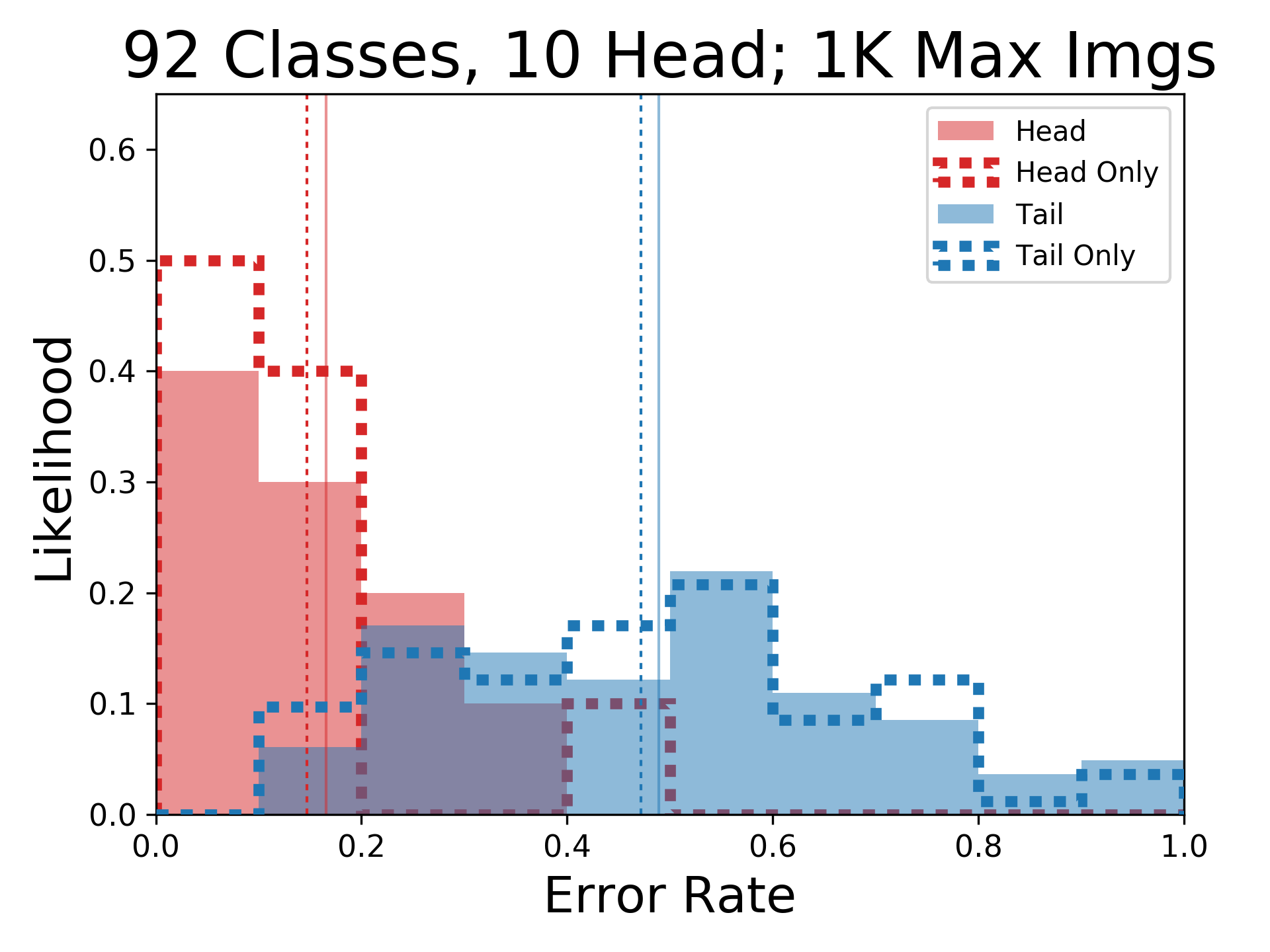}}
\subfigure[]{\label{fig:b}\includegraphics[width=0.3\textwidth]{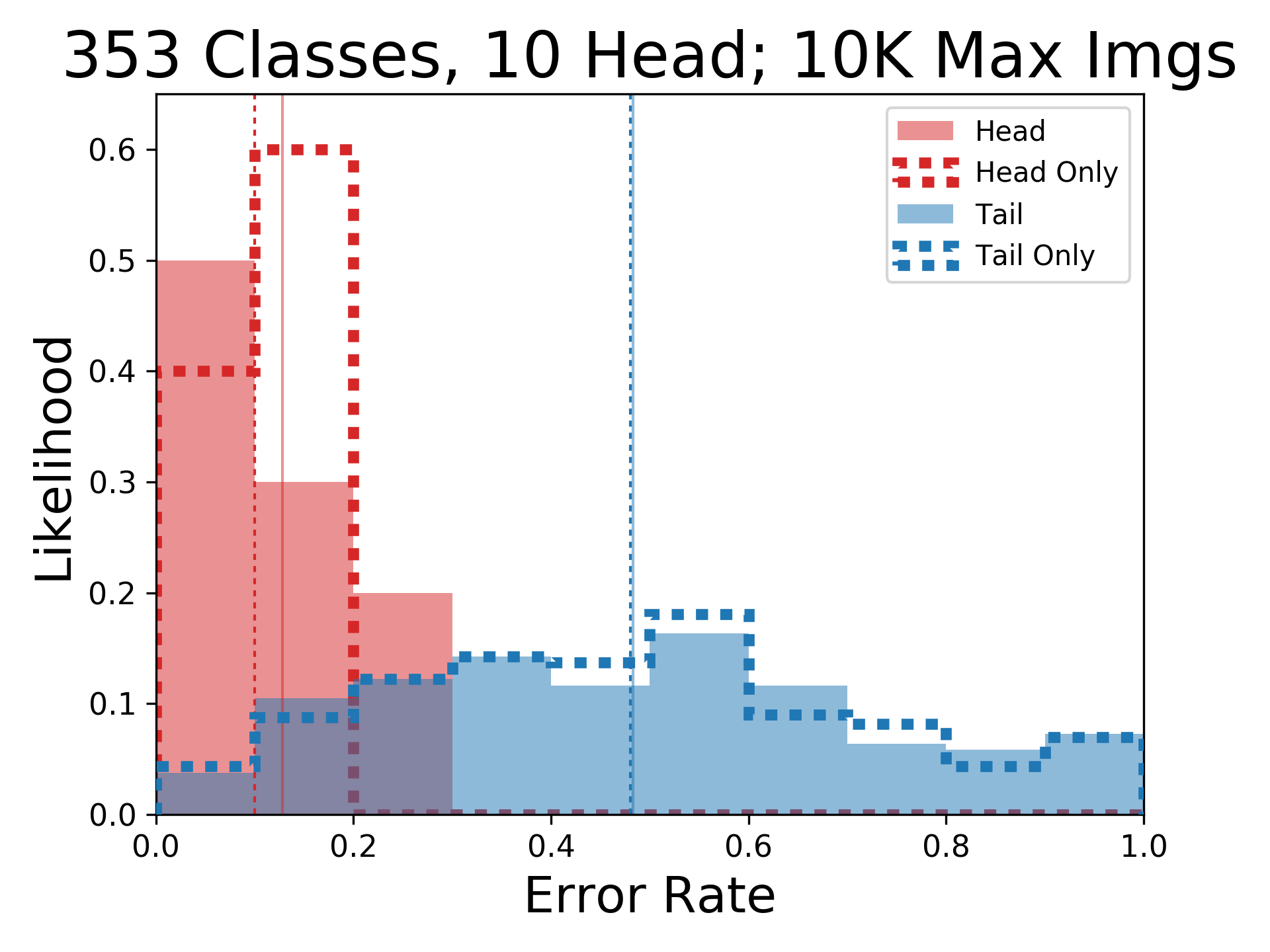}}
\subfigure[]{\label{fig:c}\includegraphics[width=0.3\textwidth]{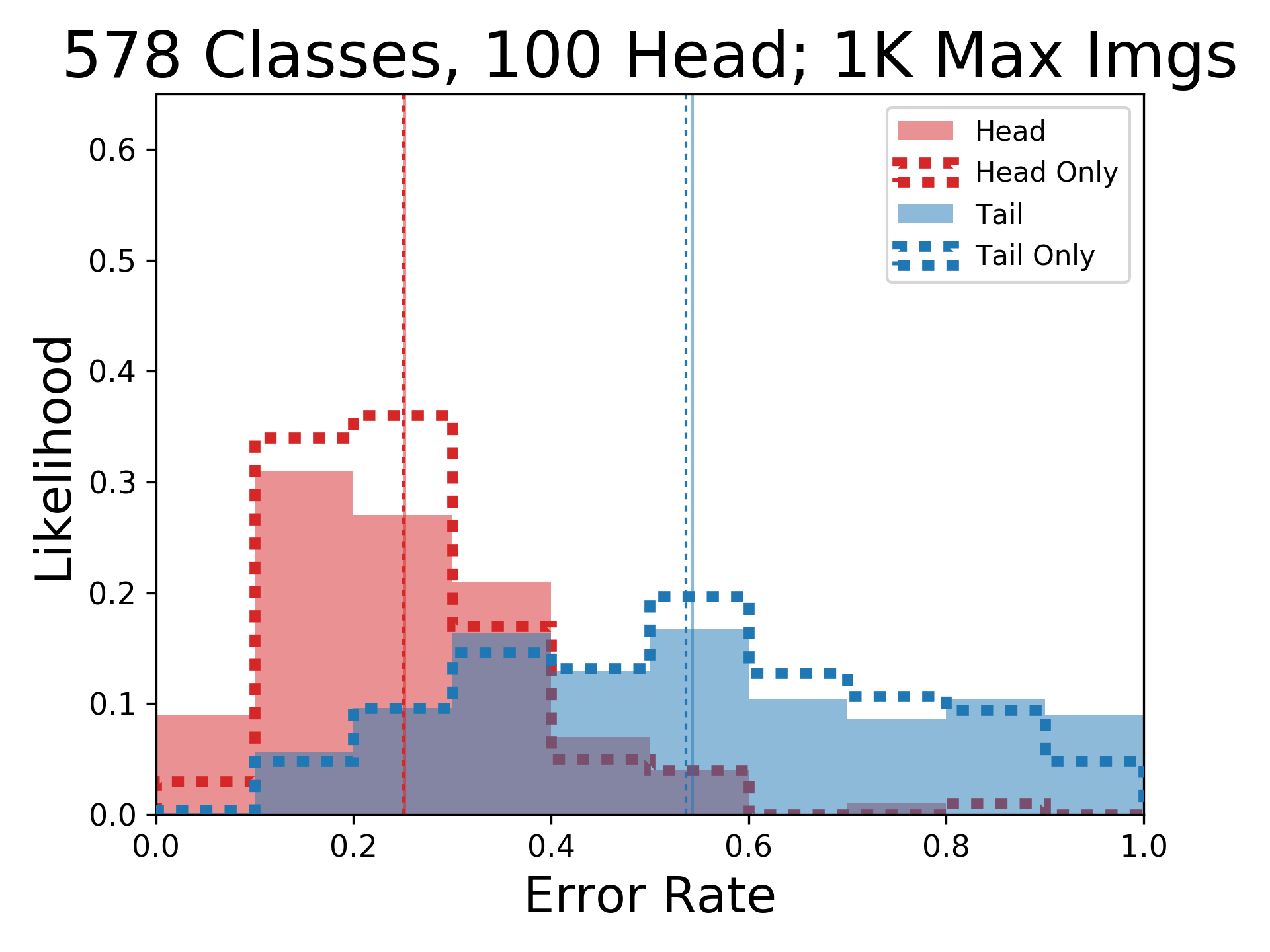}}
\Caption{\textbf{Histogram of Error Rates for Long Tail Datasets}. These plots compare the performance of the head and tail classes trained jointly (labeled Head and Tail respectively) vs individually (labeled Head Only and Tail Only respectively). The dashed histograms represent the error rates for individual models (trained exclusively on the head (red) or tail (blue) classes) and the solid histograms represent the error rates of the head and tail classes within the joint model. The vertical lines mark the mean error rates. We see that the tail classes do not benefit from being trained with the head classes: the mean error rate of a model trained exclusively on the tail classes does as good or better than a model trained with both head and tail classes. The overall joint error of the models (dominated by the tail performance) are: 45.1\% for (a), 47.1\% for (b) and 49.2\% for (c).}
\label{fig:histogram_of_error_rates}
\end{figure}

To confirm the results on the approximate long tail datasets we experimented on four long tail distributions modeled after the actual eBird dataset, see Section \ref{sec:dataset_desc} for details on the datasets. For these experiments we trained three separate models, one trained with all classes and the other two trained with the head classes or tail classes respectively. Figures \ref{fig:tail_vs_head_acc_long_tail} and \ref{fig:histogram_of_error_rates} show the results. We see the same recurring story: the tail performance is not affected by the head classes. Training a model solely on the tail classes is as good, or even better, than training jointly with the head classes, even though the head classes are from the same domain and are doubling the size of the training dataset. The network is not transferring knowledge from the head classes to the tail classes.

See the supplementary material for exact performance values of the various experiments.



%% file: conclusion.tex
\section{Discussion and Conclusions}
\label{sec:conclusion}

The statistics of images in the real world is long-tailed: a few categories are highly represented and most categories are observed only rarely. This in stark contrast with the statistics of popular benchmark datasets, such as ImageNet \cite{deng2009imagenet}, COCO \cite{lin2014microsoft} and CUB200 \cite{wah2011caltech}, where the training images are evenly distributed amongst classes. 

We experimentally explored the performance of a state-of-the-art classification model on approximate and realistic long-tailed datasets. We make four observations which, we hope, will inform future research in visual classification.

First, performance is excellent, even in challenging tasks, when the number of training images exceeds  many thousands. For example, the species classification error rate is about 4\% in the eBird dataset when each species is trained with 10$^4$ images (see Fig.\ref{fig:nn_svm_error}). This is in line with the performance observed on ImageNet and COCO, where current algorithms can rival humans. 

Second, if the number of training images is sufficient, classification performance suffers only minimally from an increase in the number of classes (see Fig.\ref{fig:nn_svm_error}). This is indeed good news, as we estimate that there are tens of millions of object categories that one might eventually attempt to classify simultaneously.

Third, the number of training images is critical: classification error more than doubles every time we cut the number of training images by a factor of 10 (see Fig.\ref{fig:nn_svm_error}). This is particularly important in a long-tailed regime since the tails contain most of the categories and therefore dominate average classification performance. For example: the largest long tail dataset from our experiments contains 550,692 images and yields an average classification error of 48.6\% (see Fig.~\ref{fig:tail_vs_head_acc_long_tail}). If the same 550,692 images  were distributed uniformly amongst the 2215 classes the average error rate would be about 27\%  (see Fig. \ref{fig:nn_svm_error}). Another way to put it: Collecting the eBird dataset took a few thousand motivated birders about 1 year. Increasing its size to the point that its top 2000 species contained at least 10$^4$ images would take 100 years (see Fig. \ref{fig:real_world_long_tail}). This is a long time to wait for excellent accuracy.

Fourth, on the datasets tested, transfer learning between classes is negligible with current classification models. Simultaneously training on well-represented classes does little or nothing for the performance on those classes that are least represented. The average classification accuracy of the models will be dominated by the poor tail performance, and adding data to the head classes will not improve the situation.  

Our findings highlight the importance of continued research in transfer and low shot learning~\cite{feifeiFP04,hariharanlow,wang2016learning,wang2016small} and provide baselines for future work to compare against. When we train on uniformly distributed datasets we sweep the world's long tails under the rug  and we do not make progress in addressing this challenge. As a community we need to face up to the long-tailed challenge and start developing algorithms for image collections that mirror real-world statistics.


%% file: supplementary.tex
\section{Supplementary}
\label{sec:supplementary}

\subsection{Detailed Results from Experiments}

\begin{table}[ht]
\centering
\begin{tabular}{|l|l|l|l|l|l|}
\hline
Dataset                                                                                      & \begin{tabular}[c]{@{}l@{}}Images / \\ Head Class\end{tabular} & \begin{tabular}[c]{@{}l@{}}Images / \\ Tail Class\end{tabular} & \begin{tabular}[c]{@{}l@{}}Overall \\ ACC\end{tabular} & \begin{tabular}[c]{@{}l@{}}Head \\ ACC\end{tabular} & \begin{tabular}[c]{@{}l@{}}Tail \\ ACC\end{tabular} \\ \hline
\multirow{4}{*}{\begin{tabular}[c]{@{}l@{}}10 head classes\\ 990 tail classes\end{tabular}}  & 100                                                            & 100                                                            & 71.5                                                   & 55.7                                                & 71.6                                                \\ \cline{2-6} 
                                                                                             & 100                                                            & 10                                                             & 33.7                                                   & 61.3                                                & 33.4                                                \\ \cline{2-6} 
                                                                                             & 1,000                                                          & 10                                                             & 34.8                                                   & 89.3                                                & 34.2                                                \\ \cline{2-6} 
                                                                                             & 10,000                                                         & 10                                                             & 35.4                                                   & 94.7                                                & 34.8                                                \\ \hline
\multirow{3}{*}{\begin{tabular}[c]{@{}l@{}}100 head classes\\ 900 tail classes\end{tabular}} & 100                                                            & 100                                                            & 71.5                                                   & 65.2                                                & 72.2                                                \\ \cline{2-6} 
                                                                                             & 100                                                            & 10                                                             & 37.9                                                   & 68.9                                                & 34.5                                                \\ \cline{2-6} 
                                                                                             & 1000                                                           & 10                                                             & 43.3                                                   & 86.1                                                & 38.6                                                \\ \hline
\end{tabular}
\caption{\textbf{Top 1 accuracy for head and tail classes when going from uniform to approximate long tail image distribution.} The uniform dataset performance is the first row for the respective datasets, the subsequent rows are approximate long tail datasets. We see that the head classes benefit from the additional training images (Head ACC increases), but the tail classes benefit little, if any (Tail ACC). }
\label{tab:uniform_to_longtail}
\end{table}

\begin{table}[ht]
\centering
\begin{tabular}{|l|l|l|l|l|l|l|l|}
\hline
Dataset                                                                & \begin{tabular}[c]{@{}l@{}}Images /\\ Head\end{tabular} & \begin{tabular}[c]{@{}l@{}}Images /\\ Tail\end{tabular} & \begin{tabular}[c]{@{}l@{}}Overall\\ ACC\end{tabular} & \begin{tabular}[c]{@{}l@{}}Head\\ ACC\end{tabular} & \begin{tabular}[c]{@{}l@{}}Tail\\ ACC\end{tabular} & \begin{tabular}[c]{@{}l@{}}Tail \\ Isolated\\ ACC\end{tabular} & \begin{tabular}[c]{@{}l@{}}$\Delta$ \\ Error\\ Tail \\ Isolated\end{tabular} \\ \hline
\multirow{7}{*}{\begin{tabular}[c]{@{}l@{}}10 H\\ 990 T\end{tabular}}  & 10                                                      & 10                                                      & 33.2                                                  & 24.7                                               & 33.2                                               & 33.4                                                           & -                                                                            \\ \cline{2-8} 
                                                                       & 100                                                     & 10                                                      & 33.7                                                  & 61.3                                               & 33.4                                               & 34.2                                                           & -1.2\%                                                                       \\ \cline{2-8} 
                                                                       & 1,000                                                   & 10                                                      & 34.8                                                  & 89.3                                               & 34.2                                               & 36.4                                                           & -4.5\%                                                                       \\ \cline{2-8} 
                                                                       & 10,000                                                  & 10                                                      & 35.4                                                  & 94.7                                               & 34.8                                               & 37.8                                                           & -6.6\%                                                                       \\ \cline{2-8} 
                                                                       & 100                                                     & 100                                                     & 71.5                                                  & 55.7                                               & 71.6                                               & 71.8                                                           & -                                                                            \\ \cline{2-8} 
                                                                       & 1,000                                                   & 100                                                     & 71.4                                                  & 83.7                                               & 71.3                                               & 71.9                                                           & -0.4\%                                                                       \\ \cline{2-8} 
                                                                       & 10,000                                                  & 100                                                     & 71.3                                                  & 94                                                 & 71                                                 & 72.6                                                           & -2.8\%                                                                       \\ \hline
\multirow{5}{*}{\begin{tabular}[c]{@{}l@{}}100 H\\ 900 T\end{tabular}} & 10                                                      & 10                                                      & 33.2                                                  & 25.8                                               & 34                                                 & 35                                                             & -                                                                            \\ \cline{2-8} 
                                                                       & 100                                                     & 10                                                      & 38                                                    & 68.9                                               & 34.5                                               & 40.5                                                           & -8.5\%                                                                       \\ \cline{2-8} 
                                                                       & 1,000                                                   & 10                                                      & 43.4                                                  & 86.1                                               & 38.6                                               & 50.9                                                           & -24.5\%                                                                      \\ \cline{2-8} 
                                                                       & 100                                                     & 100                                                     & 71.5                                                  & 65.2                                               & 72.2                                               & 73.2                                                           & -                                                                            \\ \cline{2-8} 
                                                                       & 1,000                                                   & 100                                                     & 72.8                                                  & 84.9                                               & 71.5                                               & 75.3                                                           & -7.8\%                                                                       \\ \hline
\end{tabular}
\caption{\textbf{Tail class performance.} This table details the tail class performance in uniform and approximate long tail datasets. In addition to showing the accuracy of the tail classes (Tail ACC) we show the performance of tail classes in isolation from the head classes (Tail Isolated ACC). To compute Tail Isolated ACC we remove all head class images from the test set, and ignore the head classes when making predictions on tail class images. These numbers reflect the situation of using the head classes to improve the feature representation of the network. The $\Delta$ Error Tail Isolated column shows the decrease in error between the tail performance when the head classes are considered (Tail ACC) and the tail performance in isolation (Tail Isolated ACC). These numbers are a sanity check to ensure that the tail classes do indeed benefit from a feature representation learned with the additional head class images. The problem is that the benefit of the representation is not shared when both the head and tail classes are considered together.}
\label{my-label}
\end{table}

\begin{table}[]
\centering
\begin{tabular}{|l|l|l|l|l|l|l|l|l|}
\hline
\begin{tabular}[c]{@{}l@{}}Dataset \\ Parameters\end{tabular} & \begin{tabular}[c]{@{}l@{}}Num Tail\\ Classes\end{tabular} & \begin{tabular}[c]{@{}l@{}}Overall\\ ACC\end{tabular} & \begin{tabular}[c]{@{}l@{}}Head\\ ACC\end{tabular} & \begin{tabular}[c]{@{}l@{}}Head\\ Isolated\\ ACC\end{tabular} & \begin{tabular}[c]{@{}l@{}}Head\\ Model\\ ACC\end{tabular} & \begin{tabular}[c]{@{}l@{}}Tail \\ ACC\end{tabular} & \begin{tabular}[c]{@{}l@{}}Tail\\ Isolated\\ ACC\end{tabular} & \begin{tabular}[c]{@{}l@{}}Tail\\ Model\\ ACC\end{tabular} \\ \hline
\begin{tabular}[c]{@{}l@{}}h = 10\\ y = 1K\end{tabular}       & 82                                                         & 54.9                                                  & 85.7                                               & 88.7                                                          & 87.7                                                       & 51.2                                                & 56.8                                                          & 53                                                         \\ \hline
\begin{tabular}[c]{@{}l@{}}h = 10\\ y = 10K\end{tabular}      & 343                                                        & 52.9                                                  & 89.7                                               & 94.3                                                          & 92.6                                                       & 51.8                                                & 53.6                                                          & 52.1                                                       \\ \hline
\begin{tabular}[c]{@{}l@{}}h = 100\\ y = 1K\end{tabular}      & 478                                                        & 50.8                                                  & 76.5                                               & 79.7                                                          & 76.5                                                       & 45.4                                                & 53.6                                                          & 46.1                                                       \\ \hline
\begin{tabular}[c]{@{}l@{}}h = 100\\ y = 10K\end{tabular}     & 2115                                                       & 51.4                                                  & 87.4                                               & 89.4                                                          & 87                                                         & 49.7                                                & 53.4                                                          & 53                                                         \\ \hline
\end{tabular}
\caption{\textbf{Top 1 accuracy for the long tail datasets.} This table details the results of the long tail experiments. See Section \ref{sec:dataset_desc} for information on the dataset parameters. Three different models were trained for each dataset. \textbf{1. Whole Model} This model was trained with both the head and the tail classes. Overall top 1 accuracy can be found in the Overall ACC column. Performance on the head and tail classes can be found in the Head ACC and Tail ACC columns respectively. Performance on the head and tail classes in isolation from each other can be found in the Head Isolated ACC and Tail Isolated ACC columns respectively. \textbf{2. Head Model} This model was trained exclusively on the head classes. Overall top 1 accuracy (on the head classes only) can be found in the Head Model ACC column. \textbf{3. Tail Model} This model was trained exclusively on the tail classes. Overall top 1 accuracy (on the tail classes only) can be found in the Tail Model ACC column.}
\label{my-label}
\end{table}

\clearpage

\subsection{Increasing Performance on the Head Classes}

The experiments in Section \ref{sec:tail_performance} showed that we should not expect the tail classes to benefit from additional head class training data. While we would ultimately like to have a model that performs well on the head and tail classes, for the time being we may have to be content with optimizing for the classes that have sufficient training data, \ie the head classes. In this section we explore whether we can use the tail to boost performance on the head classes. For each experiment the model is trained on all classes specified in the training regime (which may be the head classes only, or could be the head and the tail classes), but at test time only head test images are used and only the head class predictions are considered (\eg a model trained for 1000 way classification will be restricted to make predictions for the 10 head classes only).

\begin{figure}[h]
\centering
\subfigure[Uniform Datasets]{\label{fig:uniform_head_err}\includegraphics[width=0.32\textwidth]{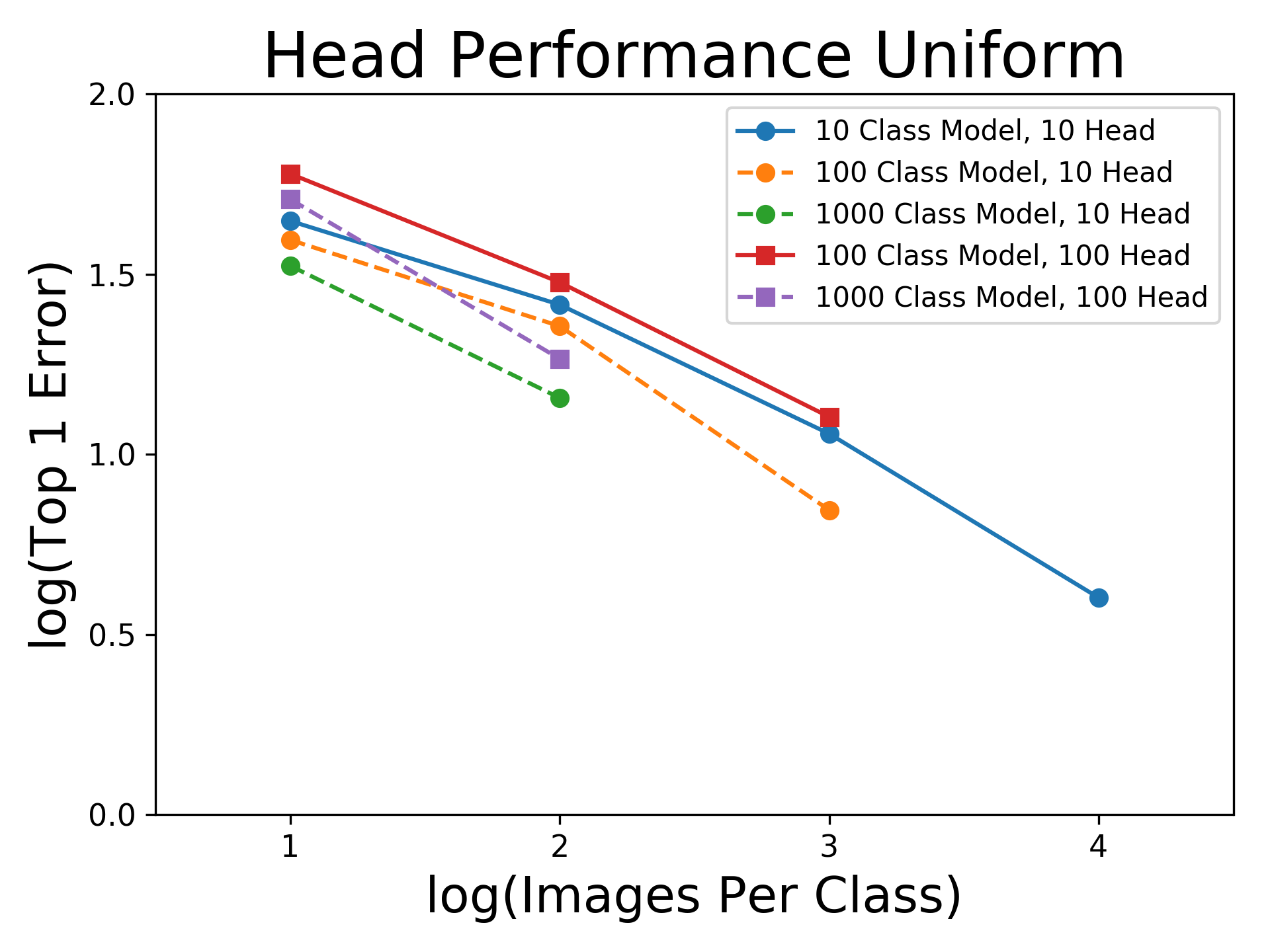}}
\subfigure[10 Head Classes, 990 Tail Classes]{\label{fig:long_tail_head_err_b}\includegraphics[width=0.32\textwidth]{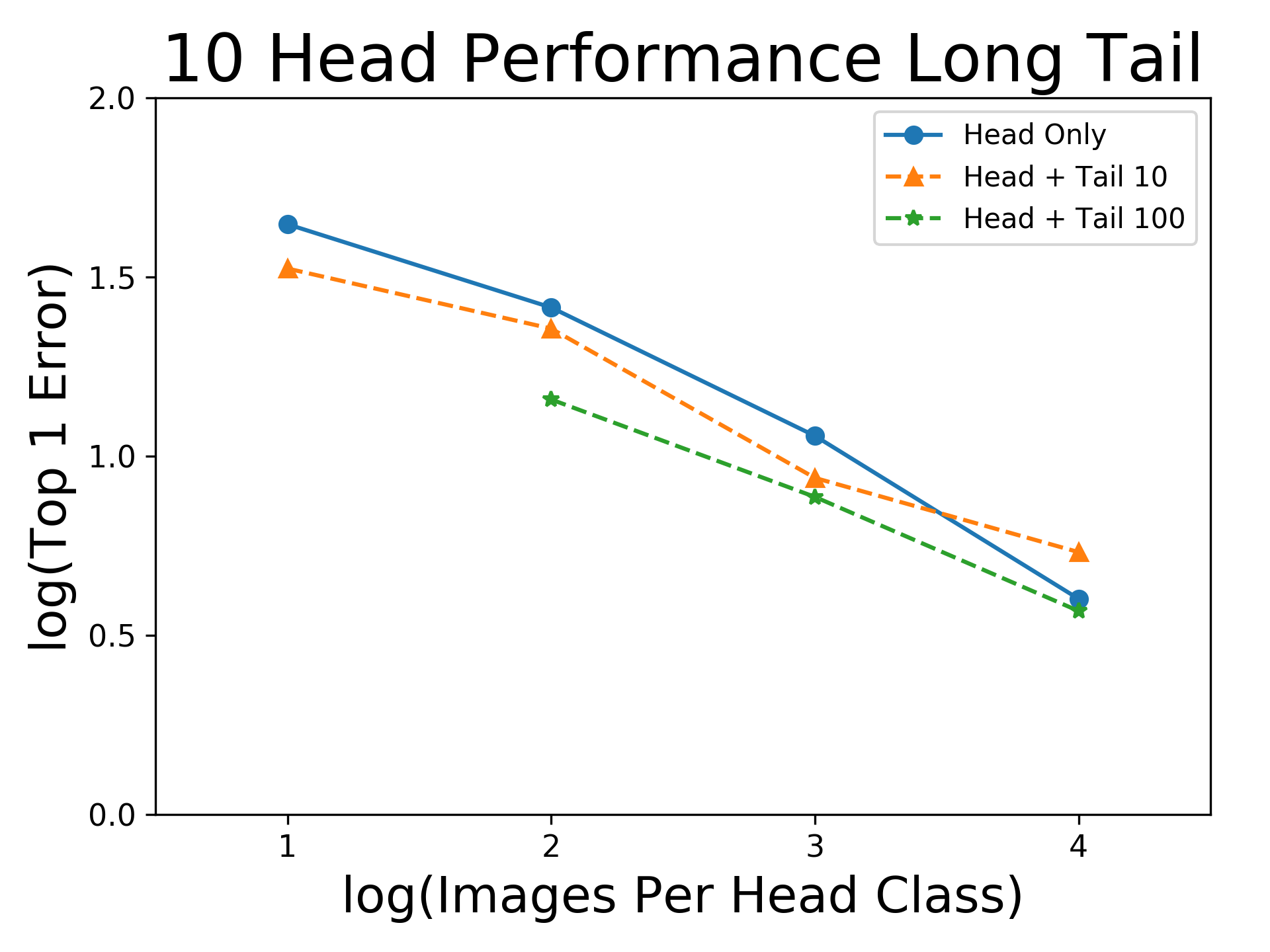}}
\subfigure[100 Head Classes, 900 Tail Classes]{\label{fig:long_tail_head_err_c}\includegraphics[width=0.32\textwidth]{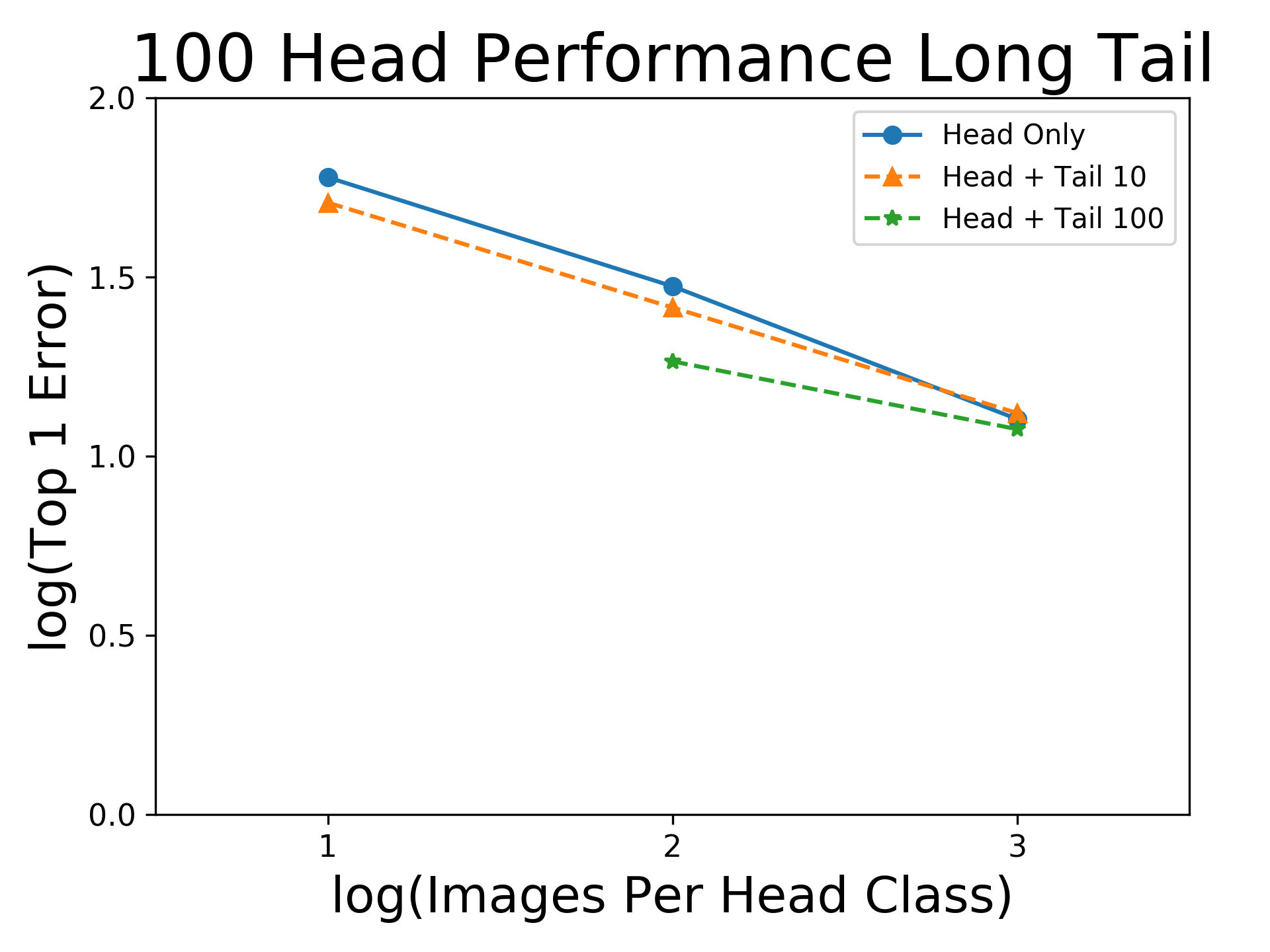}}
\caption{\textbf{Head class performance when using additional tail categories.} Head + Tail 10 refers to the tail having 10 images per class; Head + Tail 100 refers to the tail having 100 images per class. At test time we ignore tail class predictions for models trained with extra tail classes. We see that training with additional tail classes (dashed lines) decreases the error compared to a model trained exclusively on the head classes (solid lines) in both uniform and long tail datasets. In the long tail setting, the benefit is larger when the ratio of head images to tail images is smaller. We found that if this ratio exceeds 10, then it is better to train the model with the head classes only (right most points in (b) and (c)).}
\label{fig:long_tail_head_err}
\end{figure}

We first analyze the performance of the head classes in a uniform dataset situation, where we train jointly with tail classes that have the same number of training images as the head classes. This can be considered the best case scenario for transfer learning as the source and target datasets are from the same distribution and there are many more source classes than target classes. Figure \ref{fig:uniform_head_err} shows the results. In both the 10 head class situation and 100 head class situation we see drops in error when jointly training the head and tail (dashed lines) as compared to the head only model (solid lines). 

The next experiments explore the benefit to the head in the approximate long tail datasets. Figures \ref{fig:long_tail_head_err_b} and \ref{fig:long_tail_head_err_c} show the results. We found that there is a benefit to training with the long tail, between 6.3\% and 32.5\% error reduction, see Table \ref{tab:head_performance}. The benefit of the tail typically decreases as the ratio of head images to tail images increases. When this ratio exceeds 10 it is worse to use the tail during training.

\begin{figure}[h]
	\centering
    \includegraphics[width=0.45\textwidth]{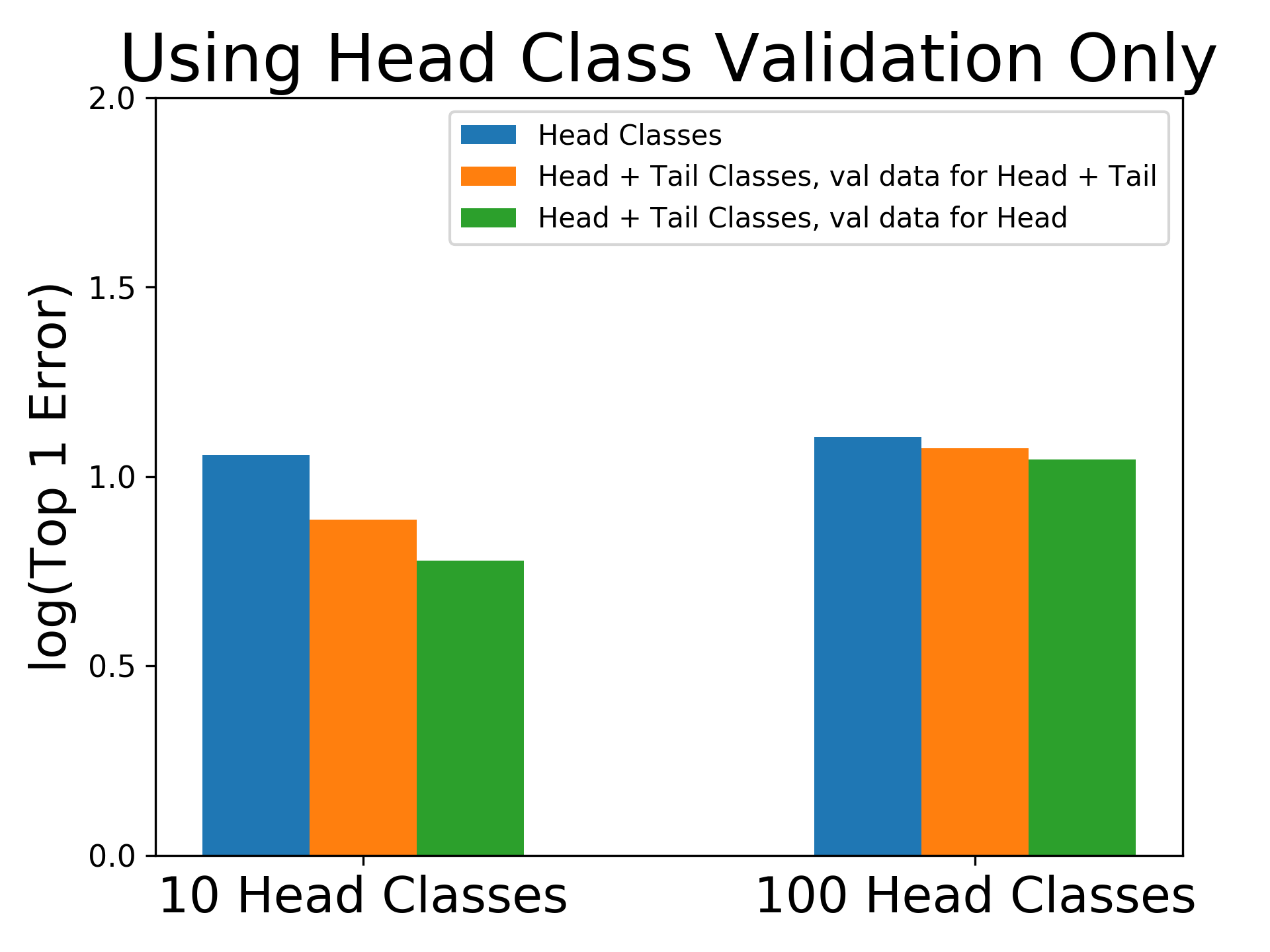}
    \caption{\textbf{Using validation data from the head classes only.} This plot shows the error achieved under different training regimes. \textbf{Head Classes} represents a model trained exclusively on the head classes, with 1000 training images each. The \textbf{Head + Tail Classes, val data for Head + Tail} represents a model trained with both head and tail classes (1000 images per head class, 100 images per tail class) and a validation set was used that had both head and tail class images. \textbf{Head + Tail Classes, val data for Head} represents a model trained with both head and tail classes (1000 images per head class, 100 images per tail class) and a validation set that only has head class images. We can see that it is beneficial to train with the extra tail classes, and that using the head classes exclusively in the validation set results in the best performing model.}
    \label{fig:head_val_only}
\end{figure}

In these experiments we have been monitoring the performance of all classes during training with a uniform validation set (10 images per class). This validation set is our probe into the model and we use it to select which iteration of the model to use for testing. We now know that using as much tail data as possible is beneficial. This raises the following question: Can we monitor solely the head classes with the validation set and still recover an accurate model? If the answer is yes then we will be able to place all tail images in the training set rather than holding some out for the validation set. The results shown in Figure \ref{fig:head_val_only} show that this is possible and that it actually produces a more accurate model for the head classes.

\begin{table}[h]
\centering
\begin{tabular}{|l|l|l|l|l|l|l|}
\hline
Dataset                                                                & \begin{tabular}[c]{@{}l@{}}Images /\\ Head\end{tabular} & \begin{tabular}[c]{@{}l@{}}Images /\\ Tail\end{tabular} & \begin{tabular}[c]{@{}l@{}}Head /\\  Tail\\ Image \\ Ratio\end{tabular} & \begin{tabular}[c]{@{}l@{}}Head\\ Isolated\\ ACC\end{tabular} & \begin{tabular}[c]{@{}l@{}}Head\\ Model\\ ACC\end{tabular} & \begin{tabular}[c]{@{}l@{}}$\Delta$ \\ Error\end{tabular} \\ \hline
\multirow{7}{*}{\begin{tabular}[c]{@{}l@{}}10 H\\ 990 T\end{tabular}}  & 10                                                      & 10                                                      & 0.01                                                                    & \textbf{66.6}                                                 & 55.6                                                       & -24.6\%                                                   \\ \cline{2-7} 
                                                                       & 100                                                     & 10                                                      & 0.1                                                                     & \textbf{77.3}                                                 & 74                                                         & -12.7\%                                                   \\ \cline{2-7} 
                                                                       & 1,000                                                   & 10                                                      & 1.01                                                                    & \textbf{91.3}                                                 & 88.6                                                       & -23.7\%                                                   \\ \cline{2-7} 
                                                                       & 10,000                                                  & 10                                                      & 10.1                                                                    & 94.6                                                          & \textbf{96}                                                & +35\%                                                     \\ \cline{2-7} 
                                                                       & 100                                                     & 100                                                     & 0.01                                                                    & \textbf{85.6}                                                 & 74                                                         & -44.6\%                                                   \\ \cline{2-7} 
                                                                       & 1,000                                                   & 100                                                     & 0.1                                                                     & \textbf{92.3}                                                 & 88.6                                                       & -32.5\%                                                   \\ \cline{2-7} 
                                                                       & 10,000                                                  & 100                                                     & 1.01                                                                    & \textbf{96.3}                                                 & 96                                                         & -7.5\%                                                    \\ \hline
\multirow{5}{*}{\begin{tabular}[c]{@{}l@{}}100 H\\ 900 T\end{tabular}} & 10                                                      & 10                                                      & 0.11                                                                    & \textbf{49}                                                   & 40                                                         & -15\%                                                     \\ \cline{2-7} 
                                                                       & 100                                                     & 10                                                      & 1.11                                                                    & \textbf{74}                                                   & 70.2                                                       & -12.8\%                                                   \\ \cline{2-7} 
                                                                       & 1,000                                                   & 10                                                      & 11.11                                                                   & 86.8                                                          & \textbf{87.3}                                              & +3.9\%                                                    \\ \cline{2-7} 
                                                                       & 100                                                     & 100                                                     & 0.11                                                                    & \textbf{81.6}                                                 & 70.2                                                       & -38.3\%                                                   \\ \cline{2-7} 
                                                                       & 1,000                                                   & 100                                                     & 1.11                                                                    & \textbf{88.1}                                                 & 87.3                                                       & -6.3\%                                                    \\ \hline
\end{tabular}
\caption{\textbf{Head class performance.} This tables details the performance of the head classes under different training regimes. The Head Isolated ACC numbers show the top 1 accuracy on the head class images when using a model trained with both head and tail classes, but only makes predictions for the head classes at test time. The Head Model ACC numbers show the top 1 accuracy for a model that was trained exclusively on the head classes. We can see that it is beneficial to train with the tail classes until the head to tail image ratio exceeds 10, at which point it is better to train with the head classes only.}
\label{tab:head_performance}
\end{table}

